\documentclass[10pt,twocolumn,letterpaper]{article}

\usepackage{iccv}
\usepackage{times}
\usepackage{epsfig}
\usepackage{graphicx}
\usepackage{amsmath}
\usepackage{amssymb}

\usepackage[pagebackref=true,breaklinks=true,colorlinks,bookmarks=false]{hyperref}

\usepackage{url}

\usepackage{tikz}
\usepackage{array}
\usepackage{anyfontsize}
\usepackage{microtype}
\usepackage{graphicx}
\usepackage{subfig}
\usepackage{booktabs} %
\usepackage{colortbl}
\usepackage{bbm}
\usepackage{bm}
\usepackage{multirow}
\usepackage{wrapfig}
\usepackage{amsthm}
\usepackage{color}
\usepackage{bbding}
\usepackage{threeparttable}
\usepackage[ruled]{algorithm2e}
\usepackage{enumitem}
\usepackage{wasysym}
\usepackage{siunitx}

\newcommand{\tol}[2]{\ensuremath{_{\pm\mathtt{#1}}^{\textcolor{blue}{\uparrow\mathtt{#2}}}}}

\newcommand{\toll}[2]{\ensuremath{_{\pm\mathtt{#1}}^{\textcolor{red}{\downarrow\mathtt{#2}}}}}
\newcommand{\tolb}[1]{\ensuremath{^{\textcolor{blue}{\uparrow\mathtt{#1}}}}}
\newcommand{\tolr}[1]{\ensuremath{^{\textcolor{red}{\downarrow\mathtt{#1}}}}}
\newcommand{\tols}[1]{\ensuremath{_{\pm\mathtt{#1}}}}

\definecolor{light-gray}{gray}{0.6}
\definecolor{light-gray-2}{gray}{0.9}

\iccvfinalcopy %

\def\iccvPaperID{11691} %

\ificcvfinal\fi

\begin{document}

\title{Towards Semi-supervised Learning with Non-random Missing Labels}

\author{Yue Duan\textsuperscript{\rm 1}
~~~Zhen Zhao\textsuperscript{\rm 2}~~~Lei Qi\textsuperscript{\rm 3}~~~Luping Zhou\textsuperscript{\rm 2}~~~Lei Wang \textsuperscript{\rm 4}~~~Yinghuan Shi\textsuperscript{\rm 1}\thanks{Corresponding author: syh@nju.edu.cn.}\\
\textsuperscript{\rm 1}Nanjing University~~~ \textsuperscript{\rm 2}University of Sydney~~~ 
\textsuperscript{\rm 3}Southeast University~~~ \textsuperscript{\rm 4}University of Wollongong\\
}

\maketitle
\ificcvfinal\thispagestyle{empty}\fi

\begin{abstract}
   Semi-supervised learning (SSL) tackles the label missing problem by enabling the effective usage of unlabeled data. While existing SSL methods focus on the traditional setting, a practical and challenging scenario called label Missing Not At Random (MNAR) is usually ignored. In MNAR, the labeled and unlabeled data fall into different class distributions resulting in biased label imputation, which deteriorates the performance of SSL models. In this work, class transition tracking based Pseudo-Rectifying Guidance (PRG) is devised for MNAR. We explore the class-level guidance information obtained by the Markov random walk, which is modeled on a dynamically created graph built over the class tracking matrix. PRG unifies the historical information of class distribution and class transitions caused by the pseudo-rectifying procedure to maintain the model's unbiased enthusiasm towards assigning pseudo-labels to all classes,
   so as the quality of pseudo-labels on both popular classes and rare classes in MNAR could be improved. Finally, we show the superior performance of PRG across a variety of MNAR scenarios, outperforming the latest SSL approaches combining bias removal solutions by a large margin. Code and model weights are available at \url{https://github.com/NJUyued/PRG4SSL-MNAR}.
\end{abstract}

\section{Introduction}
\label{sec:intro}
\begin{figure}[t]
  \centering
  \vspace{-0em}
  \includegraphics[width=0.3\textwidth]{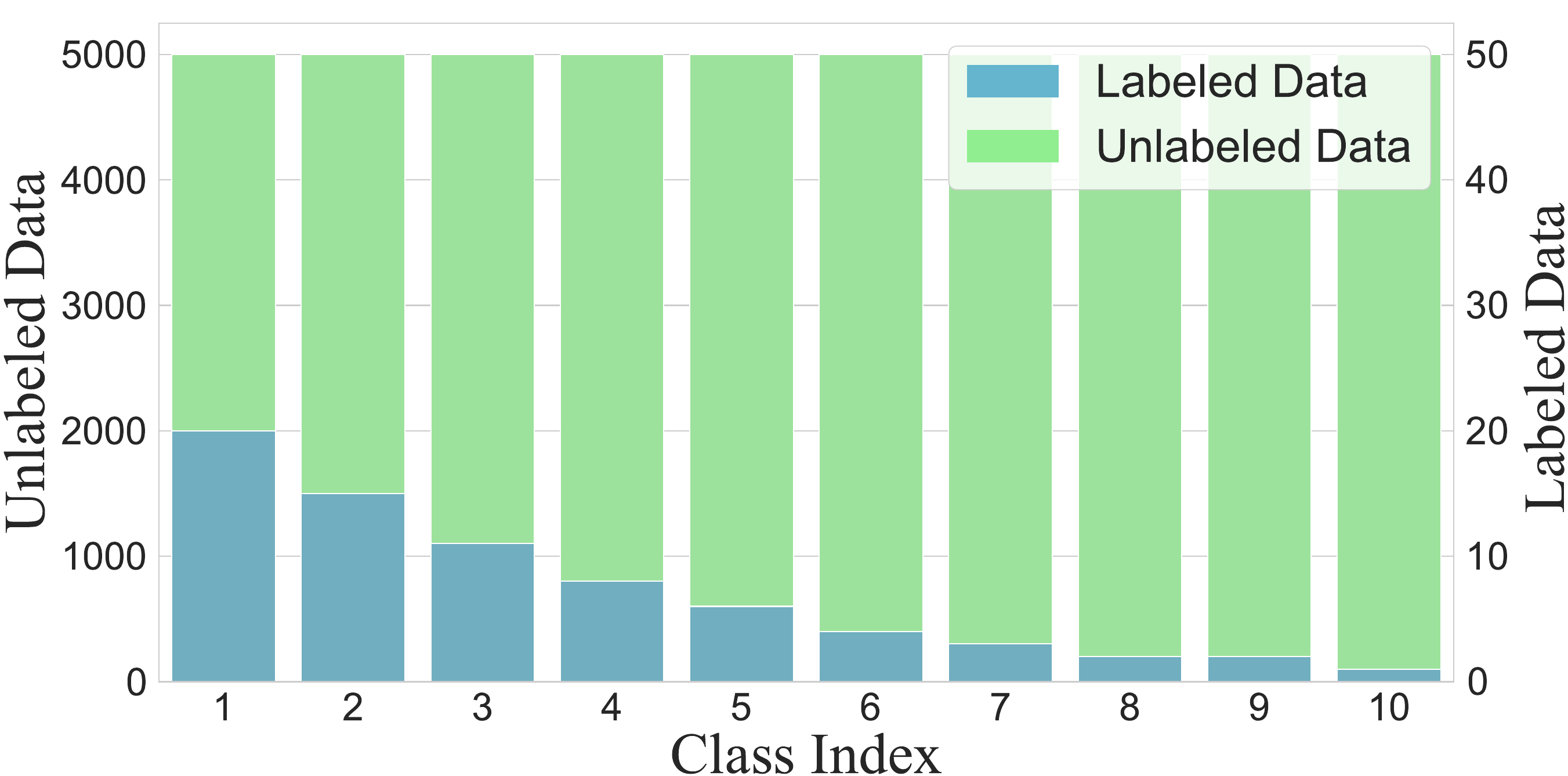}
  \caption{An example of the MNAR scenarios on CIFAR-10 (see Sec. \ref{sec:expset} for  details). The class distribution of total data is balanced whereas labeled data is unevenly distributed across classes. For better illustration, the y-axis has different scaling for labeled (\textcolor[rgb]{0.1,0.8,0.9}{blue}) and unlabeled data (\textcolor[RGB]{18,220,168}{green}).} %
    \label{fig:lu}  
\end{figure}

Semi-supervised learning (SSL), which is in the ascendant, yields promising results in solving the shortage of large-scale labeled data \cite{chapelle2009semi,van2020survey}. Current prevailing SSL methods \cite{berthelot2020remixmatch,sohn2020fixmatch,zhang2021flexmatch,duan2022mutexmatch,gui2022improving} utilize the model trained on the labeled data to impute pseudo-labels for the unlabeled data, thereby boosting the model performance. Although these methods have made exciting advances in SSL, they only work well in the conventional setting, \ie, the labeled and unlabeled data fall into the same (balanced) class distribution. Once this setting is not guaranteed, the gap between the class distributions of the labeled and unlabeled data will lead to a significant accuracy drop of the pseudo-labels, resulting in strong confirmation bias \cite{arazo2020pseudo} which ultimately corrupts the performance of SSL models. \cite{hu2021non} originally terms the scenario of the labeled and unlabeled data belonging to mismatched class distributions as label Missing Not At Random (\textbf{MNAR}) and proposes an unified doubly robust framework to train an unbiased SSL model in MNAR. During the same period, \cite{zhen2022,duan2022rda} also independently explored the issue of mismatched distributions. 
For example, a typical MNAR scenario is shown in Fig. \ref{fig:lu}, in which the popular classes of labeled data cause the model to ignore the rare classes, increasingly magnifying the bias in label imputation on the unlabeled data. It is worth noting that although some recent SSL methods \cite{kim2020distribution,wei2021crest} are proposed to deal with the class imbalance, they are still built upon the assumption of the matched class distributions between the labeled and unlabeled data, and their performance inevitably declines in MNAR.

\begin{figure*}[t]
\label{fig:2}
  \centering
    \subfloat[]{
  \includegraphics[width=4.9cm,height=3cm]{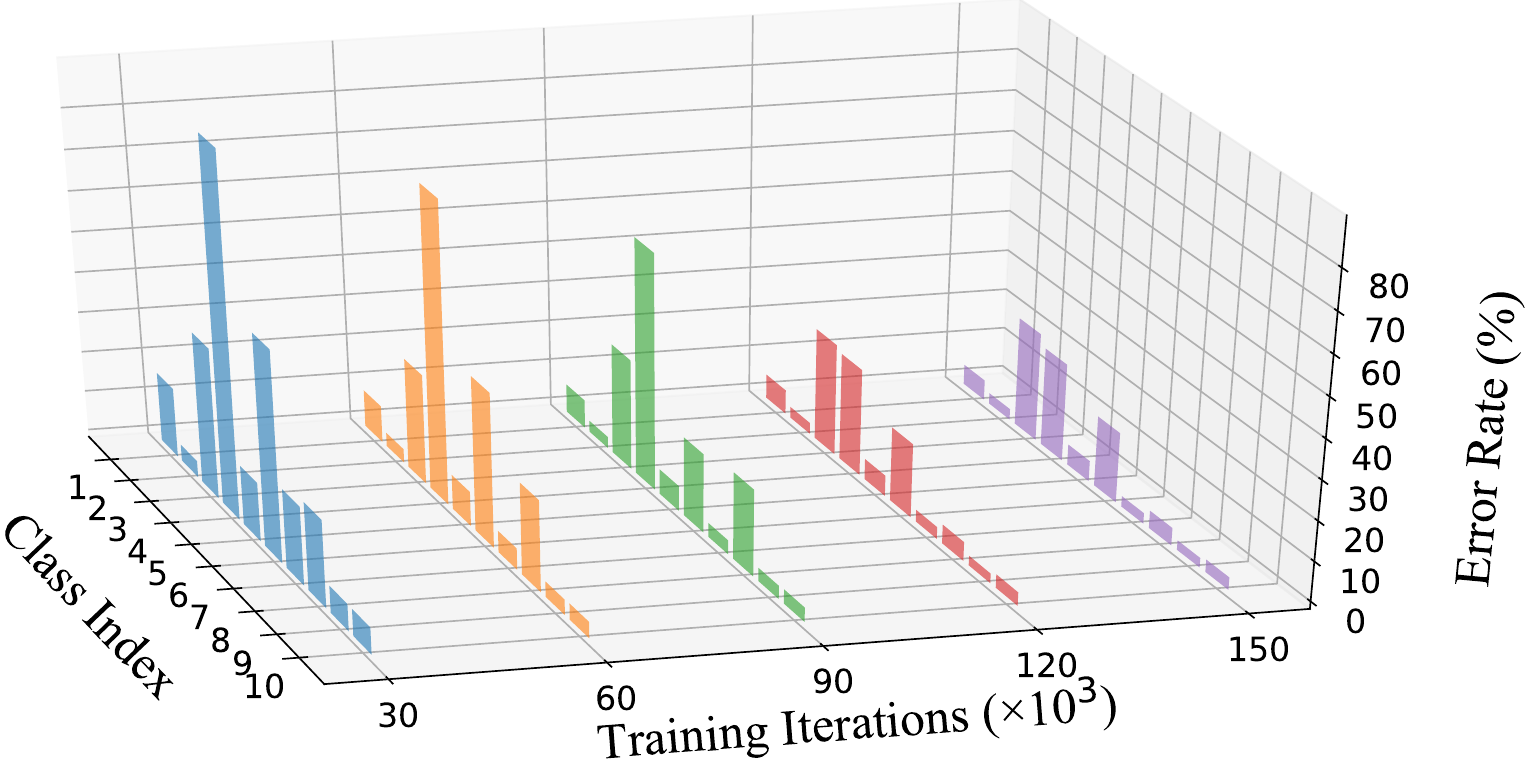}
  \label{fig:intro-c}
  }
\hspace{9mm}
  \subfloat[]{
  \includegraphics[width=4.9cm,height=3cm]{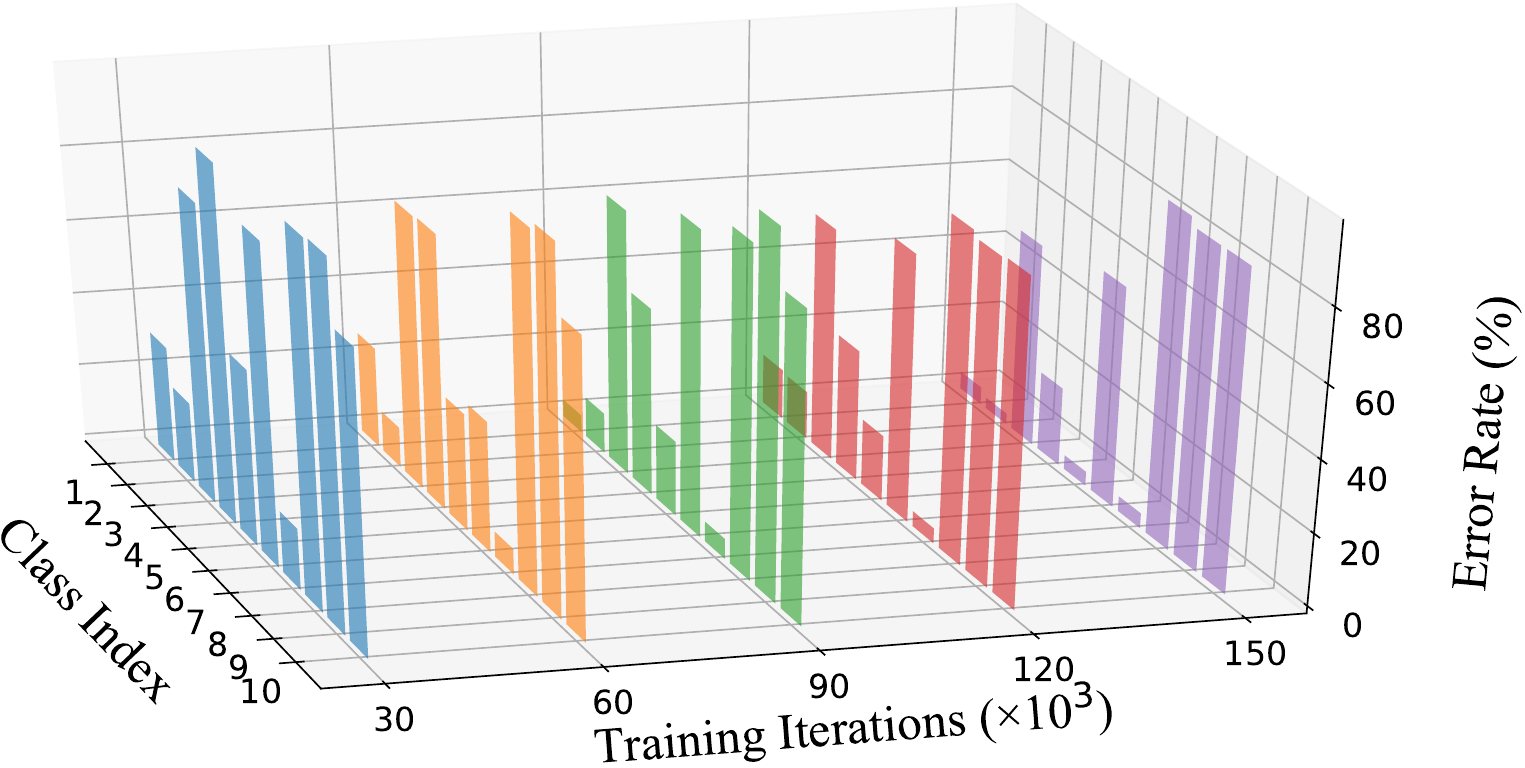}
  \label{fig:intro-b}
  }
\hspace{9mm}
  \subfloat[]{
  \includegraphics[width=3.6cm,height=3cm]{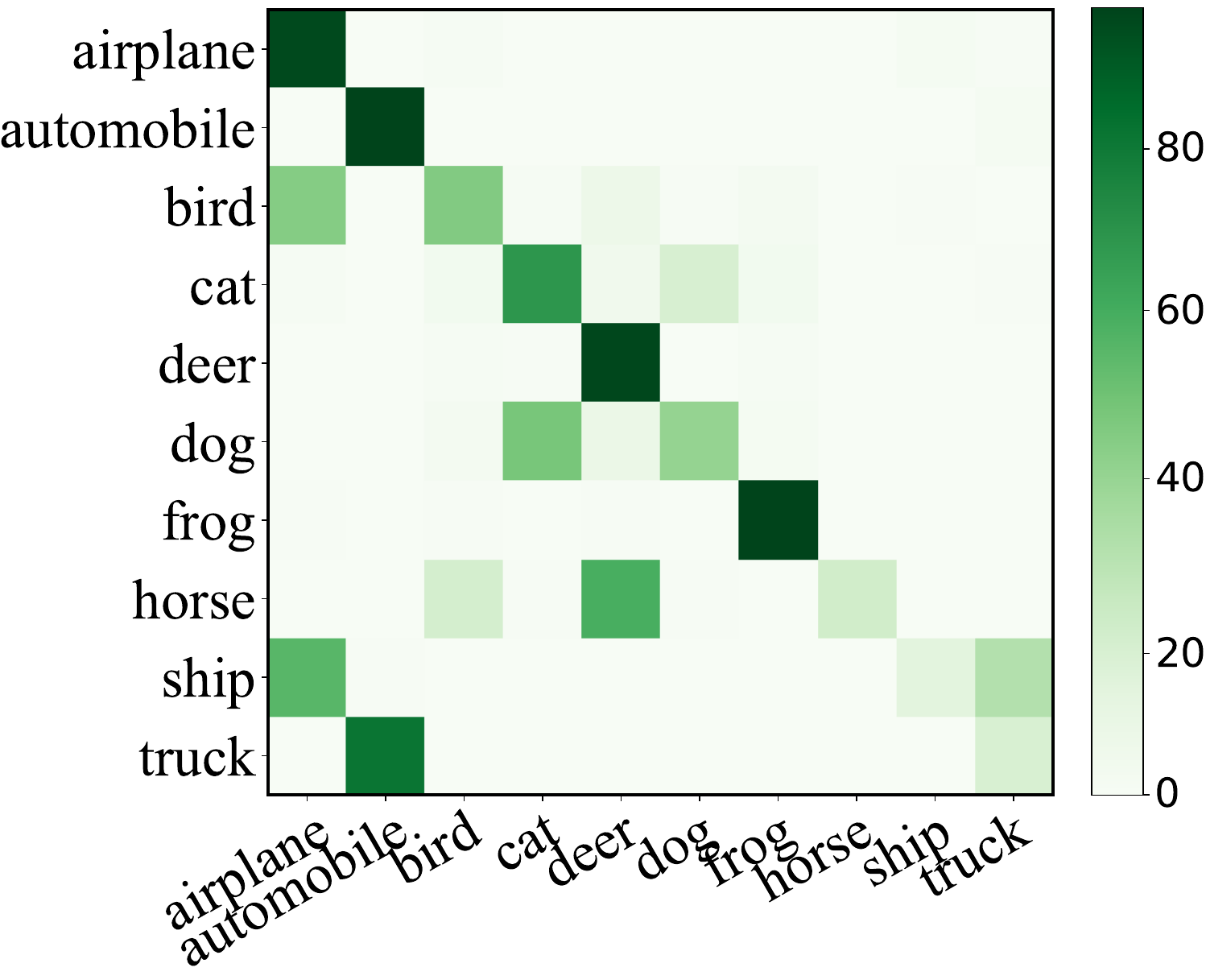}
  \label{fig:intro-a}
  }
  \caption{Results of FixMatch \cite{sohn2020fixmatch} in MNAR and the conventional SSL setting (\ie, balanced labeled and unlabeled data). The models are trained on CIFAR-10 with WRN-28-2 backbone \cite{zagoruyko2016wide}. (a) and (b): Class-wise pseudo-label error rate. (c): Confusion matrix of pseudo-labels.  In (b) and (c), experiments are conducted with the setting of Fig. \ref{fig:lu}, whereas in (a) with the conventional setting. The label amount  used in (a) is the same as that in (b) and (c). }
  \label{fig2}
  \vskip 0in
\end{figure*}

MNAR is a more realistic scenario than the 
conventional SSL setting. In the practical labeling process, labeling all classes uniformly is usually not affordable because some classes are more difficult to recognize \cite{rosset2004method,misra2016seeing}. Meanwhile, most automatic data collection methods also have difficulty in ensuring that the collected labeled data is balanced \cite{mahajan2018exploring,hu2021non}. In a nutshell, MNAR is almost inevitable in 
SSL. In MNAR, the tricky troublemaker is the mismatched class distributions between the labeled and unlabeled data. Training under MNAR, the model increasingly favors some classes,
seriously affecting the \textit{pseudo-rectifying} procedure. Pseudo-rectifying is defined as the change of the label assignment decision made by the SSL model for the same sample according to the knowledge learned at each new epoch. This process may cause \textit{class transition}, \ie, given a sample, its class prediction at the current epoch is different from that at the last epoch. In the {self-training} 
process of the SSL model driven by the labeled data, 
the model is expected to gradually rectify the pseudo-labels mispredicted for the unlabeled data in last epoches.
With pseudo-rectifying, the model trapped in the learning of extremely noisy pseudo-labels will be rescued due to its ability to correct these labels.

Unfortunately, the pseudo-rectifying ability of the SSL model could be severely perturbed in MNAR. Taking the setting in Fig. \ref{fig:lu} for example, the model's ``confidence'' in predicting the pseudo-labels into the labeled rare classes is attenuated by over-learning the samples of the labeled popular classes. Thus, the model fails to rectify those pseudo-labels mispredicted as the popular classes to the correct rare classes (even if the class distribution is balanced in unlabeled data). As shown in Fig. \ref{fig:intro-b}, compared with FixMatch \cite{sohn2020fixmatch} trained in the conventional setting (Fig. \ref{fig:intro-c}), FixMatch trained in MNAR (Fig. \ref{fig:lu})  significantly deteriorates its pseudo-rectifying ability. Even after many iterations, the error rates of the pseudo-labels predicted for labeled rare classes remain high. This phenomenon hints the necessity to provide additional guidance to the rectifying procedure to address MNAR. 
Meanwhile, as observed in Fig. \ref{fig:intro-a}, we notice that the mispredicted pseudo-labels for each class are often concentrated in a few classes, rather than scattered across all other classes. Intuitively, a class can easily be confused with the classes similar to it. For example, as shown in Fig. \ref{fig:intro-a}, the ``automobile'' samples  are massively mispredicted as the most similar class: ``truck''. Inspired by this, we argue that it is feasible to guide pseudo-rectifying from the \textit{\textbf{class level}}, \ie, pointing out the latent direction of class transition based on its current class prediction only.
For instance, given a sample classified as ``truck'', the model could be given a chance to classify it as  ``automobile'' sometimes, and vice versa. 
Notably, our approach \textit{does not} require predefined semantically similar classes. We believe that two classes are conceptually similar only if they are frequently misclassified to each other by the classifier. In this sense, we develop a novel definition of the similarity of two classes, which is directly determined by model's output. 
Even if there are no semantically similar classes, as long as the model makes incorrect prediction during the training, this still leads to class transitions which has seldom been investigated before. Our intuition could be regarded as perturbations on some confident class predictions to preserve the pseudo-rectifying ability of the model. Such a strategy does not rely on the matched class distributions assumption and therefore is amenable to MNAR.

\begin{figure*}
  \centering
  \resizebox{0.97\linewidth}{!}{   \includegraphics[]{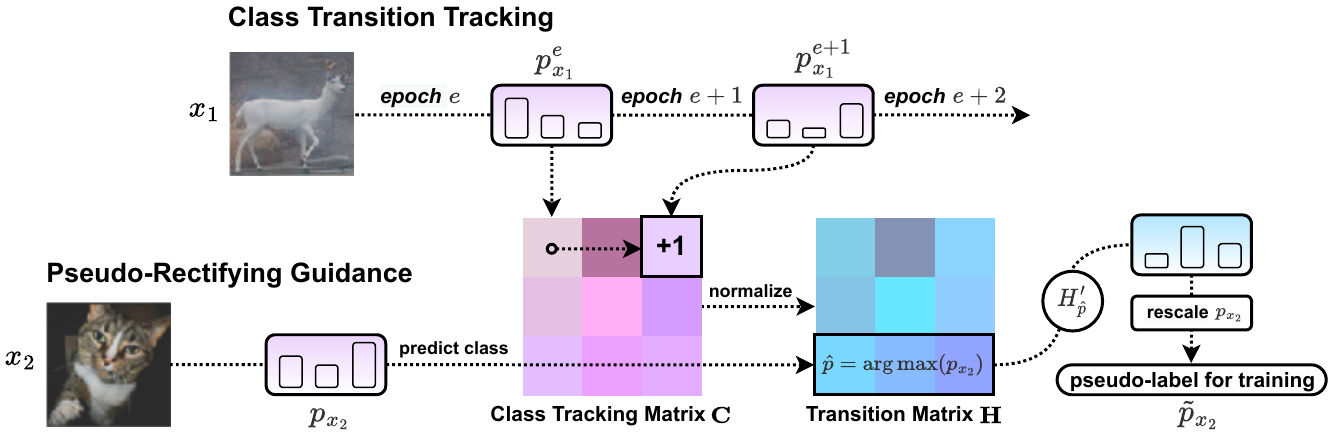}}
  \caption{Overview of PRG. Class tracking matrix $\mathbf{C}$ is obtained by tracking the  class transitions of pseudo-labels (\eg, $p_{x_{1}}$ for sample $x_{1}$) between epoch $e$ and epoch $e+1$ caused by pseudo-rectifying procedure (Eq. (\ref{eq:track})). The Markov random walk defined by transition matrix $\mathbf{H}$ (each row $H_{i}$ represents the transition probability vector corresponding to class $i$) is modeled on the graph constructed over $\mathbf{C}$. 
 Generally, given a pseudo-label, \eg, $p_{x_{2}}$ for sample $x_{2}$, class- and batch-rescaled $\mathbf{H}$ (\ie, $\mathbf{H}'$) is utilized to provide the class-level pseudo-rectifying guidance for $p_{x_{2}}$ according to its class prediction $\hat{p}=\arg\max (p_{x_{2}})$ (Eqs. (\ref{eq:sca})$\sim$(\ref{eq:de3})). Finally, the rescaled pseudo-label $\tilde{p}_{x_{2}}$ is used for the training.} 
  \label{fig:ldg}

\end{figure*}
Given the motivations above, we propose class transition tracking based Pseudo-Rectifying Guidance \textbf{(PRG)} to address SSL in MNAR, which is shown in Fig. \ref{fig:ldg}.  Our main idea can be presented as dynamically tracking the class transitions caused by pseudo-rectifying procedures at previous epoch to provide the class-level guidance for pseudo-rectifying at next epoch. 
We argue that every class transition of each pseudo-label could become the cure for the deterioration of the pseudo-rectifying ability of the traditional SSL methods in MNAR. A graph is first built on the class tracking matrix recording each pseudo-label's class transitions occurring in pseudo-rectifying procedure. 
Then we propose to model the class transition by the Markov random walk, which brings information about the difference in the propensity to rectify pseudo-labels of one class into various other classes. Specifically, we guide the class transitions of each pseudo-label during the rectifying process according to the transition probability corresponding to the current class prediction. The probability is obtained by the transition matrix of Markov random walk, and it is also rescaled based on the class distribution of assigned pseudo-labels to better provide class-level guidance. 
PRG recalls classes that are easily overlooked but appear in class transition history. They are deemed as similar to the ground-truth and have more chance to be assigned rather than simply letting the model assign the classes it favors without hesitation. In sum, PRG perturbs some confident class predictions to preserve the pseudo-rectifying ability of the model with the usage of transition history and class distribution information, which could help improve the quality of pseudo-labels suffered from biased label imputation caused by MNAR. 
We evaluate PRG on several widely-used SSL benchmarks, demonstrating its effectiveness in coping with SSL in  MNAR.

\begin{itemize}[leftmargin=*]
    \item \textbf{What is the novelty and contribution?} Towards addressing SSL in MNAR, we propose transition tracking based Pseudo-Rectifying Guidance (\textbf{PRG}) to mitigate the adverse effects of mismatched distributions via combining information from the class transition history. We propose that the pseudo-rectifying guidance can be carried out from the class level, by modeling the class transition of the pseudo-label as a Markov random walk on the graph. 
    
    \item \textbf{How about the performance improvement?}  Our solution is computation and memory friendly without introducing additional network components. PRG achieves superior performance in MNAR under various protocols and, \eg, it outperforms CADR \cite{hu2021non}, a newly-proposed method for addressing MNAR, by up to 15.11\% on CIFAR-10 and 15.21\% on mini-ImageNet in accuracy. 
\end{itemize}

\label{sec:method}

\section{Related Work}
\label{sec:related}
Addressing the issue of supervised learning requiring a large amount of labeled data has always been a focal point \cite{gu2020context,wang2022scl,li2022pln,yang2022class}. \textit{Semi-supervised learning} (SSL) is a promising paradigm to address this problem by effectively utilizing the unlabeled data. 
Given an input $x$ (labeled or unlabeled data), our objective in SSL can be described as the learning of a predictor for   generating label $y$ for it. 
In the conventional SSL \cite{lee2013pseudo,berthelot2019mixmatch,xie2020unsupervised,tai2021sinkhorn,rizve2021in,zhao2022lassl}, underlying most of them is the assumption: the distributions of labeled and unlabeled data are all balanced. 
Some more practical scenarios for SSL are now extensively discussed. 
Recently, some work has focused on addressing the class-imbalanced issue in SSL \cite{kim2020distribution,wei2021crest}. 
\cite{kim2020distribution} refines the pseudo-labels softly by formulating a convex optimization. \cite{wei2021crest} proposes class-rebalancing self-training combining \textit{distribution alignment}. 
However, these existing methods still underestimate the complexity of practical scenarios of SSL,
\eg, \cite{wei2021crest} works based on strong assumptions: the labeled data and unlabeled data fall in the same distribution (\ie, their distributions match).  
Furthermore, \cite{hu2021non,zhen2022,duan2022rda} propose a novel and realistic setting: SSL with mismatched distributions, which pops up in various fields such as social analysis, medical sciences and so on \cite{enders2010applied,heckman1977sample}. %
Typically, \cite{hu2021non} designs a class-aware doubly robust (CADR) estimator to remove the bias on label imputation caused by the mismatched distributions. Differently, our method alleviates the bias from another perspective, that is to guide the pseudo-rectifying direction based on the historical information of class transitions.

\begin{figure*}
  \centering
  \resizebox{\linewidth}{!}{   \includegraphics[]{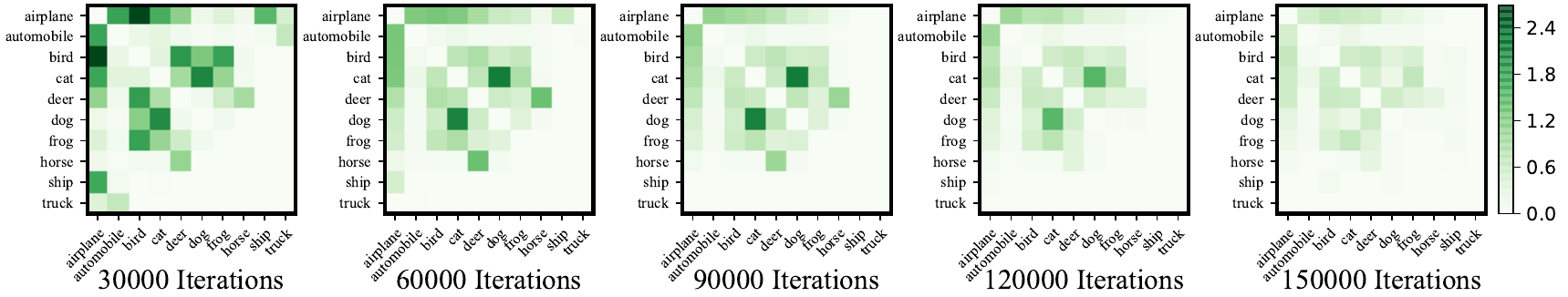}}
  \caption{Visualization of $\mathbf{C}$ obtained in training process of FixMatch \cite{sohn2020fixmatch} on CIFAR-10 with the same setting as in Figs. \ref{fig:intro-a} and \ref{fig:intro-b} (results with PRG are in Supplementary Material). The darker the color, the more frequent the class transitions. Overall, the number of class transitions decreases as the training progresses. Class transitions occur intensively between the popular classes, and class transitions between the rare classes gradually disappear (\eg, between ``ship'' and ``truck'').}
  \label{fig:met}
\end{figure*}
\section{Method}
\label{sec:met}
Formally, we denote the input space as $\mathcal{X}$ and the label space as $\mathcal{Y}=\left\{1,...,k\right\}$ over $k$ classes. SSL can be reviewed as a label missing problem, and following \cite{hu2021non}, \textit{label missing indicator} set is defined as $\mathcal{M}$ with $m\in \left\{0,1\right\}$ (only used to define the MNAR setting), where $m=1$ indicates label is missing and $m=0$ is the otherwise. Given the training dataset in SSL, we obtain a set of labeled data: $D_{L}\subseteq \mathcal{X}\times\mathcal{Y}\times\mathcal{M}$ and a set of unlabeled data:   $D_{U}\subseteq \mathcal{X}\times \widehat{\mathcal{Y}}\times\mathcal{M}$. Since the 
ground-truth $y_{U}\in \widehat{\mathcal{Y}}$ of unlabeled data $x_{U}$ is inaccessible in SSL, prevailing self-training based SSL methods impute $y_{U}$ with pseudo-label $p$. $p=f(x_{U};\theta)$ is predicted by the model  which is parametrized by $\theta$ and trained on the labeled data.  Let $(x^{(i)}_{L},y^{(i)}_{L},m^{(i)}_{L})\in D_{L}, i\in \left\{1,...,n_{L}\right\}$ be the labeled data pairs consisting of the sample with corresponding ground-truth label (\ie, $m^{(i)}=0$), and  $(x_{U}^{(i)},y_{U}^{(i)},m^{(i)}_{U})\in D_{U}, i\in \left\{n_{L}+1,...,n_{T}\right\}$ be the unlabeled data missing labels (\ie, $m^{(i)}=1$), where $n_{L}$ and $n_{T}$ refer to the number of labeled data and total training data respectively. Hereafter, the SSL dataset can be defined as $D=D_{L}\cup D_{U}$.
and we can review the conventional SSL as
a optimization task for loss $\mathcal{L}$:
\begin{equation}
  \mathop{\min}_{\theta} \sum_{(x,y,m)\in D} \mathcal{L}(x,y;\theta), 
  \label{eq:obj}
\end{equation}
where $D$ is a dataset with independent $\mathcal{Y}$ and $\mathcal{M}$. In this sense, the model trained on $D_{L}$ can easily impute unbiased pseudo-labels for unlabeled data $x_{U}$ \cite{hu2021non}. Conversely, the scenario where $\mathcal{M}$ is dependent with $\mathcal{Y}$, namely label Missing Not At Random (MNAR), will make the model produce strong bias on label imputation, which causes the ability of pseudo-rectifying suffer greatly.
Take the current most popular SSL method FixMatch \cite{sohn2020fixmatch} as an example. In FixMatch, the term $\mathcal{L}(x,y;\theta)$ in Eq. (\ref{eq:obj}) can be decomposed into two loss terms $\mathcal{L}_{L}$ and $\mathcal{L}_{U}$ with a confidence threshold $\tau$ 
\begin{align}
    \mathcal{L}(x,y;\theta)&=
    \mathcal{L}_{L}(x_{L},y_{L};\theta)+ \nonumber \\
    &\lambda_{U}\mathbbm{1}(\max(p)\geq\tau)\mathcal{L}_{U}(x_{U},\arg\max (p);\theta),
\end{align}
where $\lambda_{L}$ is the unlabeled loss weight and $\mathbbm{1}(\cdot)$ is the indicator function. Training with MNAR setting in Fig. \ref{fig:lu}, %
FixMatch is gradually seduced by samples predicted to be the labeled popular classes with confidence above $\tau$ (even though most of them are wrong), while samples predicted to be the rare class with confidence below $\tau$ do not participate into training, resulting in %
biased propensity on label imputation.
In this work, we propose class transition tracking based Pseudo-Rectifying Guidance (\textbf{PRG}) to help model better self-correct pseudo-labels with additional guidance.

\begin{algorithm*}[h]
    \small
      \caption{PRG: Class Transition Tracking Based \textbf{P}seudo-\textbf{R}ectifying \textbf{G}uidance} %
      \LinesNumbered 
      \label{a}
      \KwIn{class tracking matrices $\mathcal{C}=\{ \mathbf{C}^{(i)};i\in (1,...,N_{B})\}$, labeled training dataset $D_{L}$, unlabeled training dataset $D_{U}$, model $\theta$, label bank $\{l^{(i)};i\in (1,...,n_{T}-n_{L})\}$}
      \For{$n=1$ \rm{\textbf{to}} $\mathrm{MaxIteration}$}{
      From $D_{L}$, draw a mini-batch $\mathcal{B}_{L}=\{(x^{(b)}_{L},y^{(b)}_{L});b\in (1,...,B)\}$\\
      From $D_{U}$, draw a mini-batch $\mathcal{B}_{U}=\{(x^{(b)}_{U});b\in (1,...,B_{U})\}$\\
      
  $\mathbf{H}=\mathrm{RowWiseNormalize}(\mathrm{Average}(\mathcal{C}))$ \hfill \tcp{Construct transition matrix} 
    $H_{ij}'= \frac{H_{ij}}{\frac{L_{j}}{\sum^{k}_{d=1}L_{d}}}$ \hfill\tcp{Rescale $\mathbf{H}$ at class-level}
  \For{$b=1$ \rm{\textbf{to}} $B_{U}$}{
            $p^{(b)}=f_{\theta}(x_{U}^{(b)})$ \hfill\tcp{Compute model prediction}
            $\mathrm{idx}=\mathrm{Index}(x^{(b)}_{U})$ \hfill\tcp{Obtain the index of $x^{(b)}_{U}$ in $D_{U}$}
            $\hat{p}^{(b)}=\arg\max(p^{(b)})$ \hfill\tcp{Compute class prediction}
            \If{$l^{(\mathrm{idx})}\neq \hat{p}^{(b)}$}{ $C^{(n)}_{l^{(\mathrm{idx})}\hat{p}^{(b)}}=C^{(n)}_{l^{(\mathrm{idx})}\hat{p}^{(b)}}+1$ \hfill\tcp{Perform class transition tracking}
            $l^{(\mathrm{idx})}=\hat{p}^{(b)}$}
            $\tilde{p}^{(b)} =\mathrm{Normalize}( H_{\hat{p}^{(b)}}' \circ p^{(b)})$ \hfill\tcp{Perform pseudo-rectifying guidance}
            
  }
  $\mathcal{L}_{L},\mathcal{L}_{U}=\mathrm{FixMatch}\left(\mathcal{B}_{L},\mathcal{B}_{U},\{\tilde{p}^{(b)};b\in (1,...,B_{U})\}\right)$ \hfill\tcp{Run an applicable SSL learner}
         $\theta=\mathrm{SGD}(\mathcal{L}_{L}+\mathcal{L}_{U},\theta)$ \hfill\tcp{Update model parameters $\theta$}
      }
\end{algorithm*}

\subsection{Pseudo-Rectifying Guidance}
\label{sec:ldg}
Firstly, we formally describe the pseudo-rectifying process in SSL. In this paper, label assignment is considered as a procedure for generating soft labels. We denote the $i$-th component of vector $x$ as $x_{i}$. Let $p\in \mathbb{R}^{k}_{+}$ be the soft label vector assigned to unlabeled data $x_{U}$, where $\mathbb{R}_{+}$ is the set of nonnegative real numbers and $\sum_{i=1}^{k} p_{i}=1$. Denoting $x$ at epoch $e$ as $x^{e}$, the pseudo-rectifying process can be described as the change on $p$ by the next epoch: $p^{e+1}=g_{\theta}(p^{e})$, where $g_{\theta}(p^{e})$ is a mapping from $p^{e}$ to $p^{e+1}$ determined by the knowledge learned from the model parametrized by $\theta$ at epoch $e+1$. In MNAR, take imbalanced $D_{L}$ and balanced $D_{U}$ as an example, as the training progresses, the model's confidence is gradually slashed and unexpectedly grows on the rare and popular classes in $D_{L}$ respectively.
To address this issue, it is necessary to provide more guidance to assist the model in pseudo-rectifying.
In general, the Pseudo-Rectifying Guidance (PRG) can be described  as 
\begin{equation}
  \tilde{p}^{e+1} =\mathrm{Normalize}(\eta\circ g_{\theta}(p^{e})), 
  \label{eq:de}
\end{equation}
where $\circ$ is Hadamard product, scaling weight vector $\eta\in \mathbb{R}^{k}_{+}$ and $\mathrm{Normalize}(x)_{i}=x_{i}/{\sum_{j=1}^{k}x_{j}}$. 

We can review the technical contributions of some popular self-training works as obtaining more effective $\eta$ for pseudo-rectifying.
For example, pseudo-labeling based methods \cite{lee2013pseudo,sohn2020fixmatch,li2021comatch,xu2021dash,zhang2021flexmatch} set $\eta_{i}=1/p^{e+1}_{i}, i \in \left\{i\mid i=\arg \max(p^{e+1}) \land p^{e+1}_{i}\geq \tau\right\}$ and $\eta_{j}=0, j\in\left\{j\mid j\in (1,\cdots,k) \land j\neq i \right\}$ and, \ie, using a confidence threshold to filter low-confidence samples. 
However, it is difficult to set an apposite $\eta$ at the sample-level (\eg, for simplicity, \cite{sohn2020fixmatch} fixes $\tau$ to determine $\eta$ for all samples and the value of $\tau$ is usually set based on experience) to guide pseudo-rectifying, especially in the MNAR settings. In addition, some variants of class-balancing algorithms \cite{berthelot2020remixmatch,li2021comatch,gong2021alphamatch} can be integrated into pseudo-rectifying framework. These methods  utilize \textit{distribution alignment}  to make the class distribution of predictions close to the prior distribution (\eg, the distribution of labeled data). This process can be summarized as dataset-level pseudo-rectifying guidance by setting $\eta$ as the ratio of the current class distribution of predictions to the prior distribution, \ie, the fixed $\eta$ are used for all samples. Performing pseudo-rectifying guidance in this way strongly relies on an ideal assumptions: the labeled data and unlabeled data share the same class distribution, \ie,  in $D$, $\mathcal{Y}$ is independent with $\mathcal{M}$. Thus, these approaches fail miserably in the MNAR scenarios, which can be demonstrated in Sec. \ref{sec:dainm} of Supplementary Material.
As we discussed in Sec. \ref{sec:intro}, it is also feasible to guide pseudo-rectifying at class level. Hence, we define rectifying  weight matrix as $\mathbf{A}\in\mathbb{R}^{k\times k}_{+} $, where each row $A_{i}$ is representing the rectifying  weight vector corresponding to class $i$. Denoting the class prediction as $\hat{p}=\arg\max(p)$, the class-level pseudo-rectifying guidance can be conducted by plugging $A_{\hat{p}^{e+1}}$ into $\eta$ in Eq. (\ref{eq:de}):
\begin{equation}
  \tilde{p}^{e+1} =\mathrm{Normalize}( A_{\hat{p}^{e+1}}\circ g_{\theta}(p^{e})).
  \label{eq:de2}
\end{equation}
Next, we will introduce a simple and feasible way to obtain an effective $\mathbf{A}$ for PRG to improve the pseudo-labels. 

\begin{figure*}[t]
  \centering
  \subfloat[FixMatch]{
  \includegraphics[width=4.7cm,height=3cm]{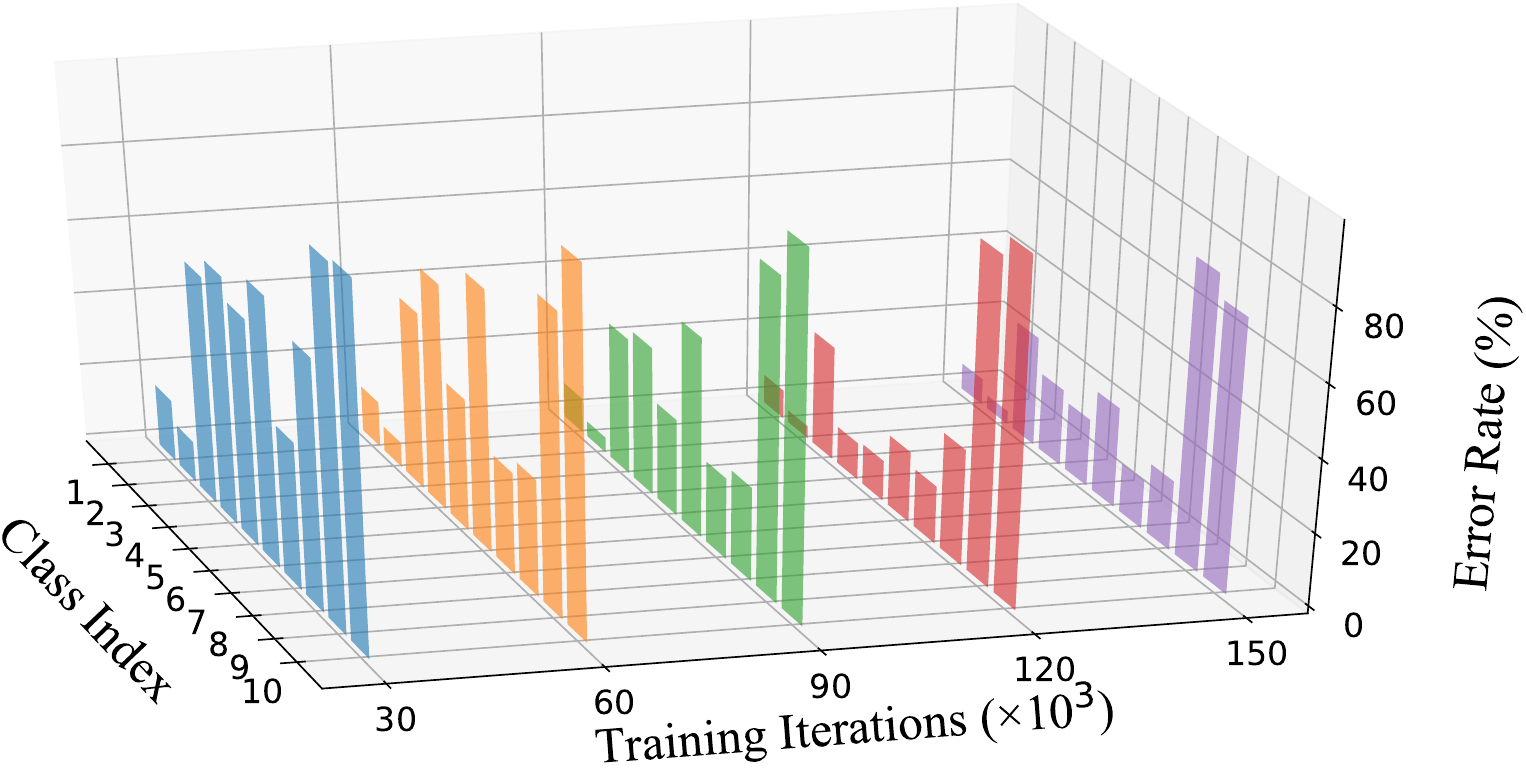}
  \label{fig:exp-a}
  }
  \hspace{9mm}
  \subfloat[PRG]{
  \includegraphics[width=4.7cm,height=3cm]{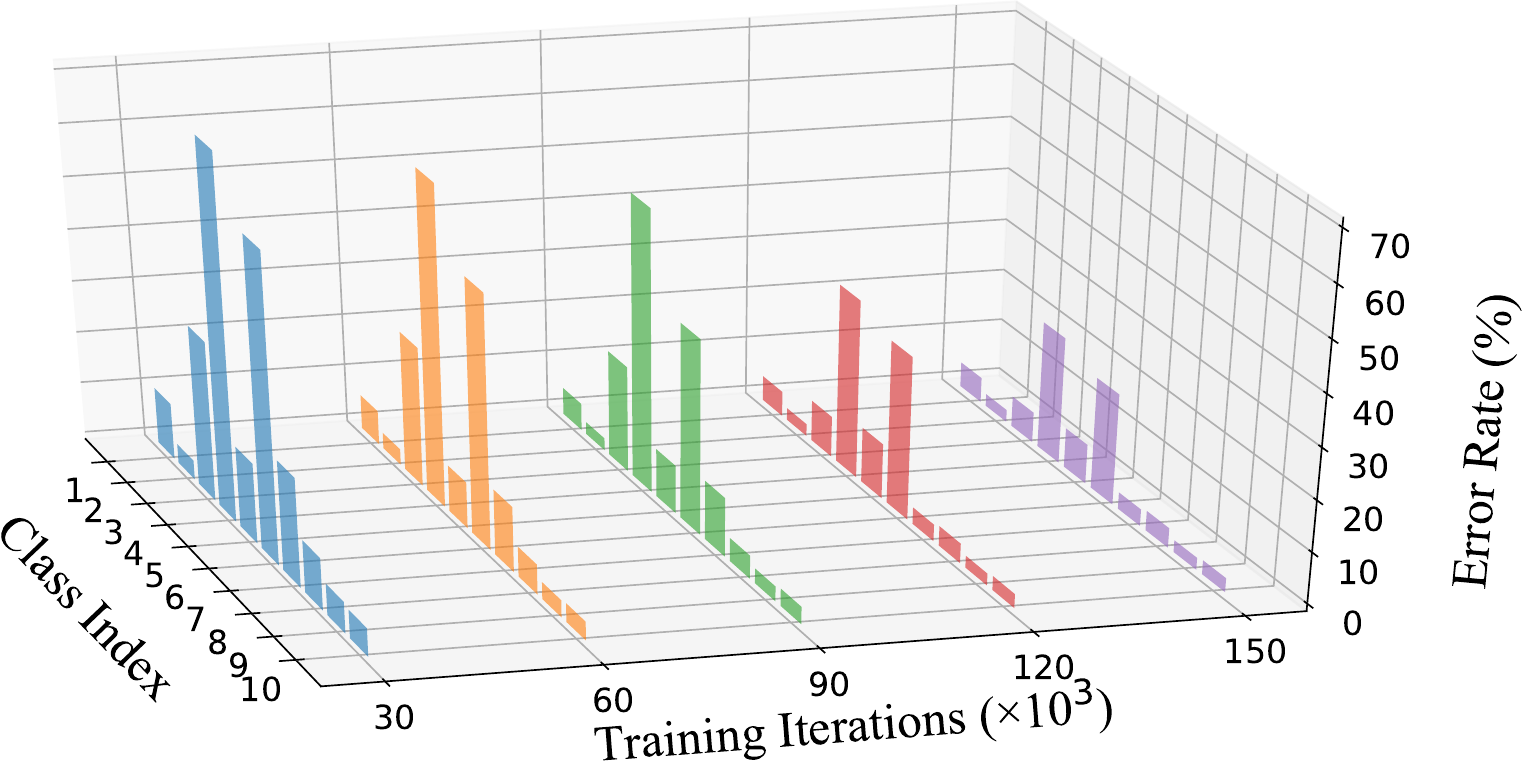}
  \label{fig:exp-b}
  }
  \hspace{9mm}
  \subfloat[PRG vs. FixMatch]{
  \includegraphics[width=4cm,height=3cm]{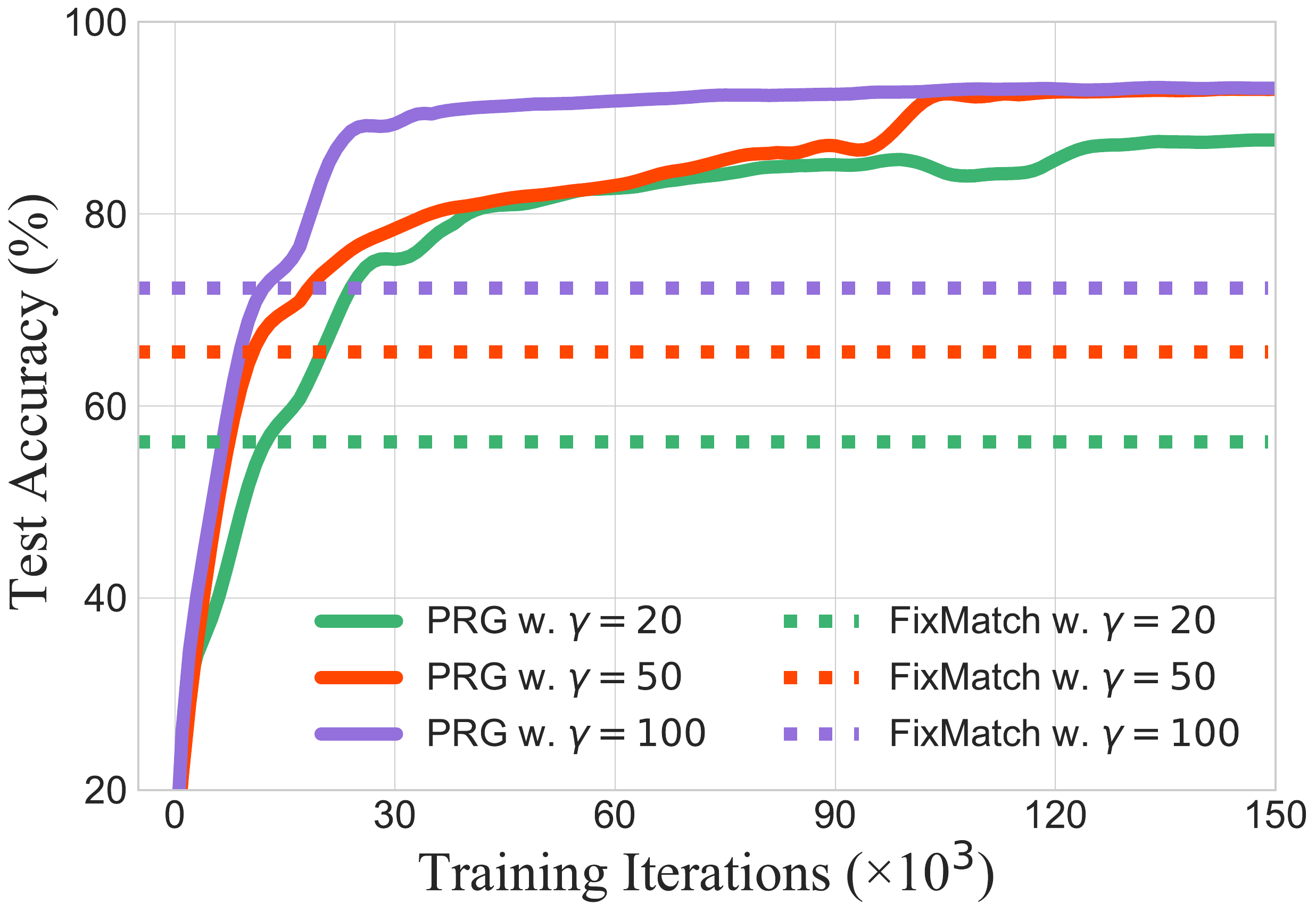}
  \label{fig:exp-c}
  }
  \caption{Results on CIFAR-10 under CADR's protocol. 
  (a) and (b): Class-wise pseudo-label error rate with $\gamma=50$. (c): Learning curve of PRG. We mark the final results of FixMatch as dash lines.}
  \label{fig:exp}
  \vskip 0in
\end{figure*}
\begin{table*}[t]
  \footnotesize
  \caption{Accuracy (\%) in  MNAR under CADR's protocol. 
The results of $^\ast$ are derived from  CADR \cite{hu2021non}. 
The larger $\gamma$, the more imbalanced the labeled data. 
  Our accuracies  are averaged on 3 runs while the standard deviations ($\pm\mathtt{Std.}$) and the performance difference (\textcolor{blue}{$\uparrow$}\textcolor{red}{$\downarrow$}$\mathtt{Diff.}$) compared to original baseline methods are reported (\textbf{bold} indicates the best results).
  }
  \label{tab:cadr}
  \vskip -0em
    \centering
    \setlength{\tabcolsep}{2.9mm}{

      \begin{tabular}{@{}l|ccc|ccc|cc@{}}     
      \toprule
      \multirow{3}{*}{Method} & \multicolumn{3}{c|}{CIFAR-10} & \multicolumn{3}{c|}{CIFAR-100} & \multicolumn{2}{c}{\quad mini-ImageNet}  \\  \cmidrule(lr){2-4}  \cmidrule(lr){5-7}  \cmidrule(l){8-9}  
   
      & $\gamma$ = 20            & 50                       & 100 & 50 &  100 & 200      & 50 &100   \\ \cmidrule(r){1-1} \cmidrule(lr){2-4}  \cmidrule(lr){5-7}  \cmidrule(l){8-9}     
      $\Pi$ Model$^\ast$    & 21.59 &27.54 &30.39 &24.95 &29.93 &33.91& 11.77 &15.30 \\
       MixMatch$^\ast$  &26.63&31.28&28.02&37.82&41.32&42.92&13.12&18.30\\
      
      ReMixMatch$^\ast$    & 41.84&38.44&38.20&42.45&39.71&39.22&22.64&23.50 \\ \cmidrule(r){1-1} \cmidrule(lr){2-4}  \cmidrule(lr){5-7}  \cmidrule(l){8-9} 
      FixMatch$^\ast$    & 56.26&65.61&72.28&50.51&48.82&50.62&23.56&26.57 \\
      + Crest$^\ast$  & 51.10\tolr{5.16}  &55.40\tolr{10.21}  &63.60\tolr{8.68} &40.30\tolr{10.21} &46.30\tolr{2.52}&49.60\tolr{1.02}&–&–\\
      + DARP$^\ast$    & 63.14\tolb{6.88}  &70.44\tolb{4.83}  &74.74\tolb{2.46}&38.87\tolr{11.64}&40.49\tolr{8.33}&44.15\tolr{6.47}&–&– \\ 
      + CADR$^\ast$    &79.63\tolb{23.37} &93.79\tolb{23.37} &93.97\tolb{21.69}&{59.53\tolb{9.02}}&60.88\tolb{12.06}&{63.30\tolb{12.68}}&29.07\tolb{5.51}&32.78\tolb{6.21} \\ 
      + PRG (Ours)   & \textbf{94.04\tol{0.18}{37.78}}
      & \textbf{94.09\tol{0.18}{28.48}}   
      & {94.28\tol{0.22}{22.00}}      
      &  59.11\tol{0.54}{8.60}
      & {61.84\tol{0.45}{13.02}}         
      & {63.41\tol{4.08}{12.79}}     
      & {44.28\tol{0.54}{20.72}}  
      & {44.99\tol{1.25}{18.42}} \\
       + PRG$^{\mathrm{Last}}$  (Ours)   & {93.81\tol{0.98}{37.55}}    & {93.44}\tol{1.05}{27.83}    & {93.48\tol{0.79}{21.20}}     &  {59.54\tol{0.99}{9.03}}     & {62.36\tol{0.23}{13.54}}          & {60.56\tol{1.86}{9.94}}     & {40.73\tol{1.27}{17.17}}   & 43.89\tol{0.14}{17.32}   \\  \cmidrule(r){1-1}\cmidrule(lr){2-4}  \cmidrule(lr){5-7}  \cmidrule(l){8-9} 
      SimMatch    & 83.45\tols{2.32}  &86.77\tols{2.15}  & 90.12\tols{1.90}  & 60.06\tols{1.17}  & 60.35\tols{0.59}  & 61.14\tols{0.24}  & 39.49\tols{1.04}  & 40.37\tols{0.96}  \\ 
       + PRG (Ours)   & {86.87\tol{2.38}{3.42}}
      & {91.68\tol{1.59}{4.91}}   
      & \textbf{94.59\tol{0.32}{4.47}}      
      &  \textbf{65.65\tol{0.61}{5.59}}
      & \textbf{65.89\tol{0.30}{5.54}}         
      & {66.50\tol{0.17}{5.36}}     
      & \textbf{44.61\tol{0.24}{5.12}}  
      & \textbf{46.48\tol{0.15}{6.11}} \\
       + PRG$^{\mathrm{Last}}$  (Ours)   & {86.46\tol{1.75}{3.01}}    & {90.48}\tol{0.96}{3.71}    & {94.22\tol{0.14}{4.10}}     &  {65.10\tol{0.40}{5.04}}     & {65.52\tol{0.29}{5.17}}          & \textbf{66.62\tol{0.19}{5.48}}     & {42.06\tol{1.81}{2.57}}   & 44.86\tol{4.49}{9.95}   \\ 
   
      \bottomrule
      \end{tabular}
  }
\end{table*}

\subsection{Class Transition Tracking}
\label{sec:ctt}
Firstly, we consider building a fully connected graph $G$ in class space $\mathcal{Y}$. This graph is constructed by adjacency matrix  $\mathbf{C}\in\mathbb{R}^{k\times k}_{+} $ (dubbed as class tracking matrix), where each element $C_{ij}$ represents the frequency of class transitions that occur from class $i$ to class $j$ (\ie, an edge directed from vertex $i$ to vertex $j$ on $G$). $C_{ij}$ is parametrized by \textit{class transition tracking} on last $N_{B}$ batches with unlabeled data batch size $B_{U}$:  $C_{ij}=\sum_{n=1}^{N_{B}}{C^{(n)}_{ij}}/N_{B}$, \ie,
\begin{align}
  & C^{(n)}_{ij}= \nonumber \\
  & \left| \left\{\hat{p}^{(b)}\mid\hat{p}^{(b),e}=i,\hat{p}^{(b),e+1}=j,i\neq j, b\in  \left\{1,...,B_{U}\right\} \right\} \right|,
    \label{eq:track}
\end{align}
where $n\in  \left\{1,...,N_{B}\right\}$ and  $C^{(n)}_{ii}=0$. Hereafter, we define the Markov random walk along the nodes of $G$, which is characterized by its transition matrix $\mathbf{H}\in\mathbb{R}^{k\times k}_{+}$ and $H_{ij}$ represents the transition probability for the class prediction $\hat{p}$ transits from class $i$ at epoch $e$  to class $j$ at epoch $e+1$. $\mathbf{H}$ is computed by conducting  row-wise normalization 
on $\mathbf{C}$.
The above designs are desirable for the following reasons.

(1) In the self-training process of the model, the historical information of pseudo-rectifying contains the relationship between classes, which is often ignored in previous methods and can be utilized to help the model assign labels at a new epoch. We can record the class transition trend in pseudo-rectifying by Eq. (\ref{eq:track}), which corresponds to the transition probability represented by $H_{ij}$, \ie, for a sample $x$, when its class prediction $\hat{p}$ is in the state of class $i$, if a rectifying procedure resulting in a class transition occurs, what probability will it transit to class $j$. Intuitively, given $p$ with $\hat{p}=i$, the model prefers to rectify it to another class similar to class $i$ in one class transition,
\ie, the preference of class transitions can also be regarded as the similarity between classes and the more similar two classes are, the more likely they are to be misclassified as each other's classes. The label is more likely to oscillate between the two classes, resulting in more swinging class transitions. As shown in Fig. \ref{fig:met}, in the ``dog'' class predictions, the predictions transitioning to the ``cat'' class are significantly more than to other classes, and vice versa in the ``cat'' labels. We can observe that $\mathbf{C}$ behaves like a symmetric matrix, reflecting the symmetric nature of class similarity.
Consequently, this similarity between classes can be utilized to provide information for our class-level pseudo-rectifying guidance. 

(2) In the MNAR settings, the tricky problem is that the mismatched distributions lead to biased label imputation for unlabeled data. 
The feedback loop of self-reinforcing errors is not achieved overnight. Empirically, as the training progresses, the model becomes more and more confident in the popular classes (in labeled or unlabeled data), which leads to misclassify the samples that it initially thought might be the rare classes  to the popular classes later. As shown in Fig. \ref{fig:met}, the lower left corner and upper right corner of the heatmap (\ie, the class transitions between the popular classes and rare classes) is getting lighter and always lighter than the upper left corner (\ie, the class transitions among the popular classes), which means the model is increasingly reluctant to transfer the class prediction to the rare classes during the pseudo-rectifying  process. 
If we only focus on what the model has learned at present, the model's past efforts to recognize the rare classes will be buried. The latent relational information between classes is hidden in the pseudo-rectifying process producing class transitions. The history of class transitions can point the way for bias removal on label imputation with an abnormal propensity on different classes caused by mismatched distributions in MNAR.

\begin{table*}[t]
  \footnotesize
  \caption{Accuracy (\%) in MNAR under our protocol with the varying labeled data sizes $n_{L}$ and imbalanced ratios $N_{1}$. Baseline methods are based on our reimplementation.
  }
  \label{tab:dc}
  \vskip -0em
    \centering
    \setlength{\tabcolsep}{2.5mm}{

      \begin{tabular}{@{}l|cccc|cc|cc@{}}     
      \toprule
      \multirow{3}{*}{Method}& \multicolumn{2}{c}{CIFAR-10 ($n_{L}$ = 40)} & \multicolumn{2}{c|}{CIFAR-10 ($n_{L}$ = 250)} & \multicolumn{2}{c|}{CIFAR-100 ($n_{L}$ = 2500)} & \multicolumn{2}{c}{mini-ImageNet ($n_{L}$ = 1000)}  \\  \cmidrule(lr){2-3} \cmidrule(lr){4-5}  \cmidrule(lr){6-7}  \cmidrule(lr){8-9} 
      & $N_{1}$ = 10            & 20                       & 100 & 200 & ~~~ 100 & 200      & ~~~40& \!\!\!\!80   \\ \cmidrule(r){1-1}\cmidrule(lr){2-3} \cmidrule(lr){4-5}  \cmidrule(lr){6-7}  \cmidrule(lr){8-9}    
      FixMatch    & 85.72\tols{0.93} &76.53\tols{3.03}    & 69.76\tols{5.57}&46.53\tols{8.12}&~~~{61.31\tols{3.67}}&41.38\tols{2.84}&~~~36.20\tols{0.36}&\!\!\!\!28.33\tols{0.41} \\
    + CADR   & {85.54\toll{2.07}{0.18}}      & {75.11\tol{3.41}{1.42}} \hspace{0.1em}   & {92.25\tol{1.61}{22.49}}      &  {63.92\tol{5.47}{17.39}}    & ~~~\textbf{61.62\tol{0.93}{0.31}}           & {46.16\tol{1.45}{4.78}} \hspace{0.1em}    & ~~~{36.08\toll{0.84}{0.12}}   & \!\!\!\!{30.52\tol{0.99}{2.19}}  \\
      + PRG (Ours)    & \textbf{91.87\tol{1.05}{6.15}}    & {77.44\tol{15.96}{0.91}}      & \textbf{93.93\tol{0.16}{24.17}}     &  \textbf{67.86\tol{16.98}{21.33}}      & ~~~{61.49\tol{3.93}{0.18}}             & \textbf{49.84\tol{1.37}{8.46}}        & ~~~\textbf{39.99\tol{0.76}{3.79}}      & \!\!\!\!\textbf{35.39\tol{0.47}{7.06}}      \\ %
      + PRG$^{\mathrm{Last}}$ (Ours)    & {85.66\toll{5.93}{0.06}}      & \textbf{77.85\tol{1.86}{1.32}} \hspace{0.1em}   & {92.80\tol{1.44}{23.04}}      &  {64.00\tol{5.02}{17.47}}    & ~~~{60.41\toll{1.01}{0.90}}           & {43.80\tol{1.71}{2.42}} \hspace{0.1em}    & ~~~{39.84\tol{0.05}{3.64}}   & \!\!\!\!{33.17\tol{0.52}{4.84}}  \\
      \bottomrule
      \end{tabular}
  
  }
\end{table*}

With obtained $\mathbf{H}$, 
some preparations are done for plugging it into  Eq. (\ref{eq:de2}) to replace $\mathbf{A}$. We're only modeling the pseudo-rectifying process resulting in class transition (\ie, $C_{ii}=0$), which means $H_{ii}=0$, \ie, $\eta_{\hat{p}^{e+1}}$ is set to $0$ in Eq. (\ref{eq:de}). This will encourage the class prediction to transition to other classes during each pseudo-rectifying process, which is unreasonable for training a robust classifier. Hence, we control the probability that does not transition class by setting $H_{ii}=\frac{\alpha}{k-1}$, where $\frac{1}{k-1}$ is the average of the transition probabilities in each row of $\mathbf{H}$ and $\alpha$ is a pre-defined hyper-parameter.
In addition, to better provide class-level guidance, we scale each element in $\mathbf{H}$ by 
\begin{equation}
    H_{ij}'= \frac{H_{ij}}{\frac{L_{j}}{\sum^{k}_{d=1}L_{d}}}
    \label{eq:sca}
\end{equation}
where $L\in \mathbb{R}_{+}^{k}$ and $L_{i}$ records the  number of class predictions belonging to class $i$ averaged on last $N_{B}$ batches. 
To sum up, we are trying to fight against MNAR by yielding the effect of adjusting the class distribution of pseudo-labels (\ie, $\frac{L_{c}}{\sum^{k}_{d=1}L_{d}}$). After all, the trouble MNAR brings us is biased label imputation caused by mismatched class distribution. We encourage the model to perform the pseudo-rectifying process, which leads to more labels being transition to classes with too few assigned labels, rather than ignoring the rare classes due to over-learning of popular classes.
More explanations and alternatives of Eq. (\ref{eq:sca}) can be found in Sec. \ref{sec:appd} of Supplementary Material.
Hereafter, we plug $\mathbf{H}'$ into $\mathbf{A}$ in Eq. (\ref{eq:de2}) for pseudo-rectifying guidance framework:
\begin{equation}
  \tilde{p}^{e+1} =\mathrm{Normalize}(H'_{\hat{p}^{e+1}}\circ g_{\theta}(p^{e})), 
  \label{eq:de3}
\end{equation}
where $H'_{\hat{p}^{e+1}}$ can be regarded as the class prediction for one sample randomly walks along the nodes of $C$ at the current epoch, \ie, drive a possible class transition in the pseudo-rectifying for bias removal on label imputation propensity due to MNAR. 
It is also feasible to use the class transition driven by $\hat{p}^{e}$ to revise $p^{e+1}$ (what is the class prediction after a class transition), \ie, replace $H'_{\hat{p}^{e+1}}$ in Eq. (\ref{eq:de3}) with $H'_{\hat{p}^{e}}$, which is dubbed as \textbf{PRG$\bm{^{\mathrm{Last}}}$}. The algorithms of PRG 
and PRG$\bm{^{\mathrm{Last}}}$ are presented in Algorithm \ref{a} and Algorithm \ref{a2} in Supplementary Material, respectively.

\textbf{Why does our method work for MNAR?} In the setting of MNAR, being aware of rare class plays a key role, PRG enhances the model to preserve a certain probability to generate class transition to rare classes when assigning pseudo-labels. This form of probability based on class transition history produces effective results, because we do not spare any attempt of the model to identify the rare class by class transition tracking (such attempts would be slowly buried due to overlearning of the popular classes). Thereby, PRG helps the model to still try to identify rare classes with a certain probability while  combines the class distribution information of pseudo-labels so that the model can assign labels to rare classes with a clear purpose.

\begin{figure*}[t]
  \centering
  \subfloat[Our protocol]{
  \includegraphics[width=6.65cm,height=2cm]{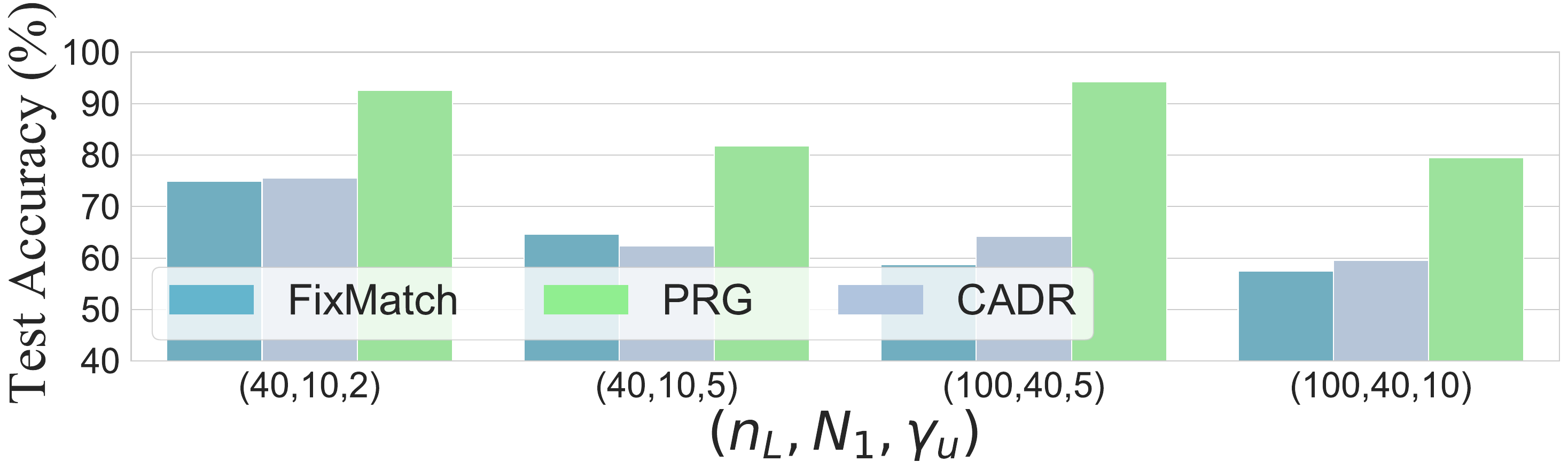}
  \label{fig:tab34-a}
  }
\hfil
  \subfloat[DARP's protocol with $\gamma_{l}=100$]{
  \includegraphics[width=6.65cm,height=2cm]{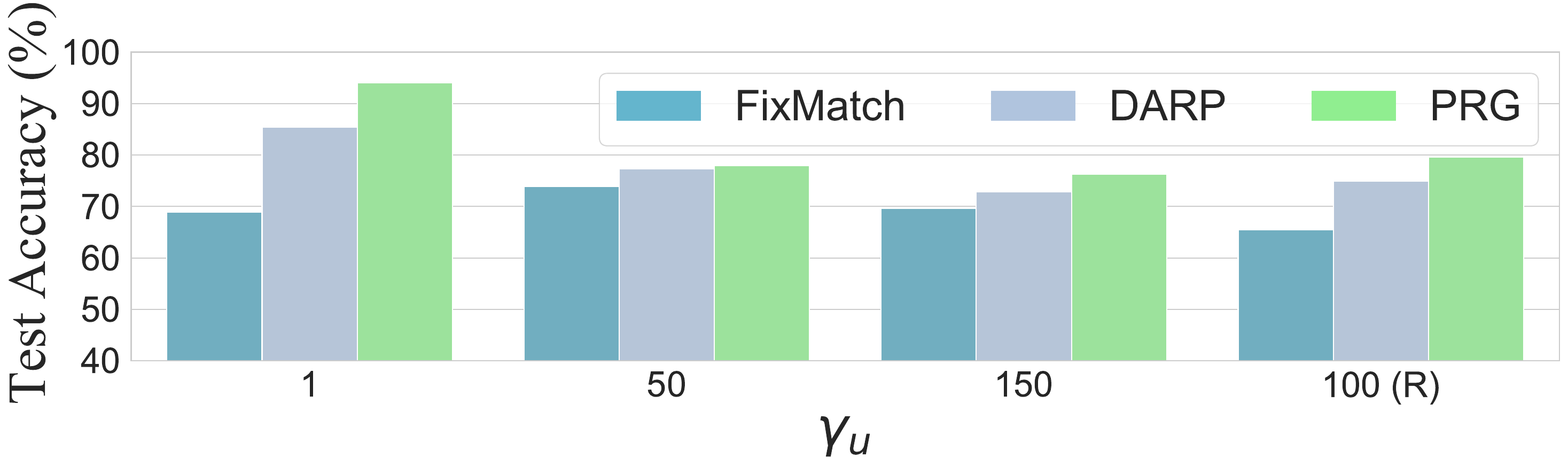}
  \label{fig:tab34-c}
  }
  \caption{Results on CIFAR-10 under two protocols.
  The imbalanced distributions of labeled and unlabeled data are in reverse order of each other in (a) and the case of $\gamma_{u}$ marked with \textit{``R''} in (b). }
  \label{fig:tab34}
  \vskip -0em
\end{figure*}

\section{Experiment}
\label{sec:exp}

\noindent\textbf{Dataset and Baselines.} We evaluate PRG  on three widely used SSL benchmarks, including CIFAR-10, CIFAR-100  \cite{krizhevsky2009learning} and mini-ImageNet \cite{vinyals2016matching} 
(a subset of ImageNet \cite{deng2009imagenet} composed of 100 classes). 
Following \cite{hu2021non}, 
we mainly report the mean accuracy of PRG in both conventional SSL settings and various MNAR scenarios. Multiple baseline methods are compared, including conventional SSL algorithms: $\Pi$ Model \cite{rasmus2015semi}, MixMatch \cite{berthelot2019mixmatch}, ReMixMatch \cite{berthelot2020remixmatch}, FixMatch \cite{sohn2020fixmatch} and SimMatch \cite{zheng2022simmatch}.
More importantly, we provide comparisons with the recent label bias removal methods for imbalanced SSL: DARP \cite{kim2020distribution}, Crest \cite{wei2021crest}, and the latest approache designed for addressing SSL in MNAR: CADR \cite{hu2021non}. 

\noindent\textbf{MNAR Settings.} 
\label{sec:expset}
Following \cite{hu2021non}, the MNAR scenarios are mimicked by constructing the class-imbalanced subset of the original dataset for either $D_{L}$ or $D_{U}$. Let $\gamma$ denote the imbalanced ratio, $N_{i}$ and $M_{i}$ respectively refer to the number of the labeled and the unlabeled data in class $i$ from $k$ classes.
Three MNAR protocols are used for the evaluations on PRG: (1) CADR's protocol \cite{hu2021non}. $N_{i}=\gamma^{\frac{k-i}{k-1}}$, in which $N_{1}=\gamma$ is the maximum number of labeled data in all classes, and the larger the value of $\gamma$, the more imbalanced $D_{L}$. For example, Fig. \ref{fig:lu} shows CIFAR-10 with $\gamma=20$. (2) Our protocol. Because the total number of labeled data $n_{L}$ in the CADR's protocol varies with $\gamma$, which violates the principle of controlling variables, $n_{L}$ is fixed by users in our protocol.
$N_{1}$ is altered for different scales of imbalance, \ie, $N_{i}=N_{1}\times \gamma^{-\frac{i-1}{k-1}}$ while $\gamma$ is calculated by the constraint $\sum_{i=1}^{k}N_{i}=n_{L}$. We further consider the MNAR settings where $D_{U}$ is also imbalanced, \ie, $M_{i}=M_{1}\times \gamma_{u}^{-\frac{k-i}{k-1}}$ (implying inversely imbalanced distribution compared with $D_{L}$), where $M_{1}=5000$ in CIFAR-10. (3) DARP'sprotocol \cite{kim2020distribution}: $N_{i}=N_{1}\times \gamma_{l}^{-\frac{i-1}{k-1}}$, $M_{i}=M_{1}\times \gamma_{u}^{-\frac{i-1}{k-1}}$, where $N_{1}=1500$ and $M_{1}=3000$ in CIFAR-10, where  $\gamma_{l}$ and $\gamma_{u}$ are varied for $D_{L}$ and $D_{U}$ respectively, \ie, the distributions of $D_{L}$ and $D_{U}$ are mismatched and imbalanced.

\noindent\textbf{Implementation Details.} In this section, PRG is mainly implemented as a plugin to FixMatch \cite{sohn2020fixmatch} and SimMatch \cite{zheng2022simmatch}. Thus, we keep the same hyper-parameters as their original paper, whereas the class invariance coefficient $\alpha=1$ and the tracked batch number $N_{B}=128$ are set for PRG. The complete list of hyper-parameters can be found in Sec. \ref{sec:id2} of Supplementary Material. Following \cite{sohn2020fixmatch}, our models are trained for $2^{20}$ iterations, respectively using the backbone of WideResNet-28-2 (WRN) \cite{zagoruyko2016wide} for CIFAR-10, WRN-28-8 for CIFAR-100 and ResNet-18 \cite{he2016deep} for mini-Imagenet.
\begin{table}[t]
\vspace{-0em}
\caption{Geometric mean scores (GM) in MNAR on CIFAR-10 under CADR's protocol.
  }
  \vspace{-0.5em}
  \label{tab:gm}
    \centering
    \footnotesize
    \setlength{\tabcolsep}{3.5mm}{
    
\begin{tabular}{@{}lccc@{}}
\toprule
Method & $\gamma=20$ &  $\gamma=50$ & $\gamma=100$    \\ \midrule
FixMatch     &  41.90\tols{8.55}&	53.61\tols{2.29}&	60.35\tols{1.84}                      \\
+ CADR     &  75.25\tol{1.55}{33.35}&	92.98\tol{0.43}{39.37}&	93.15\tol{0.36}{32.8}                      \\
+ PRG (Ours)    &\textbf{93.53\tol{0.39}{51.63}}	&\textbf{93.70\tol{0.20}{40.19}} &	\textbf{93.94\tol{0.35}{33.69}}                         \\
+ PRG$^{\mathrm{Last}}$ (Ours)     &  93.35\tol{1.10}{51.45} &	92.99\tol{1.17}{39.38} &	93.25\tol{0.97}{32.90}
               \\ \bottomrule           
\end{tabular}

}
\end{table}

\noindent\textbf{Experimental Results List (SM refers to Supplementary Material).}   (1) Imbalanced $D_{L}$ and balanced $D_{U}$ / Mismatched  imbalanced $D_{L}$ and $D_{U}$: \textit{Main Results} / \textit{More MNAR Settings} in Sec. \ref{sec:res}.
(2)  Balanced $D_{L}$ and imbalanced
$D_{U}$: Sec. \ref{app:mns} of SM. 
(3) Balanced $D_{L}$ and $D_{U}$: Sec. \ref{app:con} of SM.
(4) \textit{More Application Scenarios} in Sec. \ref{sec:res}.
(5) Ablation studies: Sec. \ref{sec: con}.
(6) Results with distribution alignment: Sec. \ref{sec:dainm}.
(7) More evaluation metrics  (\eg, precision and
recall): Sec. \ref{app:mm} of SM.
(8) PRG built on other SSL learners: Sec. \ref{app:msl} of SM.

\subsection{Results in MNAR Settings}
\label{sec:res}

\begin{table*}[t]
\caption{  Accuracy (\%) on tabular MNIST. $\gamma$ is varied for CADR's protocol whereas $n_{L}$ and $N_{1}$ are varied for our protocol. }
  \label{tab:ab11}
    \centering
    \footnotesize
    \setlength{\tabcolsep}{3mm}{
     
\begin{tabular}{@{}lccccccc@{}}
\toprule
Method & $\gamma=20$ &  $50$ & $100$ &$n_{L},N_{1}=40,10$&$40,20$&$250,100$&$250,200$    \\ \midrule
VIME&{63.38$\pm$4.42}&63.75$\pm$6.10&64.80$\pm$2.76&50.13$\pm$7.56&30.73$\pm$8.69&60.58$\pm$2.68 & 21.44$\pm$0.58             \\
+ PRG (Ours)& 59.41$\pm$14.45& {65.92$\pm$13.90} & \textbf{66.60$\pm$12.58 }& 49.28$\pm$11.09 &\textbf{34.08$\pm$16.05} &  \textbf{66.14$\pm$11.88} &  \textbf{24.51$\pm$9.56}\\
+ PRG$^{\mathrm{Last}}$ (Ours)& \textbf{63.49$\pm$10.73}& \textbf{66.19$\pm$14.22} & 66.21$\pm$10.24 & \textbf{53.17$\pm$8.84} &32.45$\pm$10.10 &  65.25$\pm$9.46 &  23.62$\pm$11.39\\\bottomrule
\end{tabular}
}
\end{table*}
\noindent\textbf{Main Results.} The experimental results under CADR's and our protocol with various levels of imbalance are summarized in Tabs. \ref{tab:cadr} and \ref{tab:dc}. PRG consistently wins baseline methods across most of the settings, benefiting from the information offered by class transition tracking. As shown in Figs. \ref{fig:exp-a} and \ref{fig:exp-b}, the pseudo-rectifying ability of PRG is significantly improved  compared with the original FixMatch, \ie, as the training progresses, the error rates of both the popular classes and the rare classes of the labeled data are greatly reduced, eventually yielding improvements in test accuracy shown in Fig. \ref{fig:exp-c}. Meanwhile, in Tab. \ref{tab:gm}
we further provide \textit{geometric mean scores} (GM, a metric often used for imbalanced dataset \cite{kubat1997addressing,kim2020distribution}), which is defined by the geometric mean over class-wise sensitivity for evaluate the classification performance of models trained in MNAR.

Our main competitors include three categories. (1) State-of-The-Art (SOTA) SSL methods:  FixMatch \cite{sohn2020fixmatch} and SimMatch \cite{zheng2022simmatch}. As shown in Tabs. \ref{tab:cadr} and \ref{tab:dc}, these methods show poor performance under MNAR. FixMatch almost can't cope with MNAR, whereas with our method, the performance is significantly improved by more than 10\% in most cases. Likewise, SimMatch's performance  is also improved by a large margin.  (2) Imbalanced SSL methods: DARP \cite{kim2020distribution} and Crest
\cite{wei2021crest}. These two SOTA methods addressing long-tailed distribution in SSL emphasize the bias removal in matched distribution (\ie, the unlabeled data is equally imbalanced as the labeled data), showing very limited capacity in handling MNAR. (3) SSL solutions devised for the MNAR scenarios: CADR \cite{hu2021non}. Our method outperforms CADR under its proposed protocol across the board, 
demonstrating PRG is more effective for bias removal on label imputation than it. With extremely few labels, the class-aware propensity estimation in CADR is not reliable whereas our method still works well, 
yielding a performance gap of up to 14.41\%.

\noindent\textbf{More MNAR Settings.}
More MNAR scenarios are considered for evaluation. In our protocol, we alter $N_{1}$ and $\gamma_{u}$ to mimic the case where the distributions of the labeled and unlabeled data are imbalanced and mismatched, \ie, the two distributions are different. Likewise, DARP's protocol produces similar mismatched distributions. As shown in Fig. \ref{fig:tab34}, PRG achieves promising results in all the comparisons with the baseline methods.  Our method boosts the accuracy of FixMatch by up to 35.51\% and 24.33\% in our and DARP's protocols respectively.
The activated class transitions make the model less prone to over-learning  unexpected classes so that the negative effect of MNAR can be mitigated.

\noindent\textbf{More Application Scenarios.}
To explore the broader image recognition applications of PRG, we further apply it to tabular MNIST (handwritten digital image from $10$ classes \cite{lecun1998gradient} interpreting as tabular data with 784 features) by plugging it into VIME \cite{yoon2020vime} (see Sec. \ref{app:mdt} of Supplementary Material for details). 
As shown in Tab. \ref{tab:ab11}, PRG outperforms baselines in the most of settings and its upper performance limits  consistently exceed VIME by a large margin. Performance fluctuations can be alleviated by adjusting $\alpha$ (here we keep consistent setting as previous experiments). 
The scheme of class-transition-based pseudo-rectifying guidance is high-level and general for classification task, so it shows promising potential in a broader MNAR scenario.

\subsection{Ablation Studies} 
\label{sec: con}
\noindent\textbf{Re-weighting scheme on $\mathbf{H}$.} 
As shown in Tab. \ref{tab:ab1}, the re-weighting scheme can effectively boost the performance of PRG in MNAR because it better provide class-level guidance by involving class distribution information and controls the intensity of class transition. Additionally, for the utilization of $\mathbf{H}'$ in Eq. (\ref{eq:de3}), we consider taking $k$ steps, \ie, multiply by $\mathbf{H}'^{k}$ instead of $\mathbf{H}'$ to uncover more complex patterns of misclassification than simple pairwise class relations. However, as shown in Tab. \ref{tab:ab2}, we can observe that the performance is inversely proportional to $k$. The advantage of PRG is that $\mathbf{H}'$ is updated in each iteration, which means that the value of $\mathbf{H}'$ is dynamic. As the model learns new knowledge, the past $\mathbf{H}'$ may not be suitable for the pseudo-rectifying process anymore. If $\mathbf{H}'^{k}$ is used, this means that we are using the same $\mathbf{H}'$ multiple times for a given sample, which wastes the advantage of dynamic $\mathbf{H}'$. $\mathbf{H}'^{k}$ using a suitable $k$ or a dynamic selection of $k$ might yield better performance, but it is complicated to determine the value of $k$. Therefore,  PRG is designed for simplicity and exhibits superior performance.

\begin{table}
      \caption{Ablation studies on re-weighting scheme in Eq. (\ref{eq:sca}). We report the results (accuracy (\%) / GM) on CIFAR-10 under CADR's protocol.
  }
  \label{tab:ab1}
    \centering
    \footnotesize
     \setlength{\tabcolsep}{2.9mm}{
\begin{tabular}{@{}lccc@{}}
\toprule
Method & $\gamma=20$ &  $\gamma=50$ & $\gamma=100$    \\ \midrule
PRG wo. Eq. (\ref{eq:sca})      &  88.97 / 87.37 &	91.73 / 91.28 &	92.72 / 92.55                     \\
PRG     & {94.04 / 93.53}	&{94.09 / 93.70	} &	{94.28 / 93.94} \\ \bottomrule
\end{tabular}
}

    \end{table}

\begin{figure}[t]
  \includegraphics[width=0.46\textwidth]{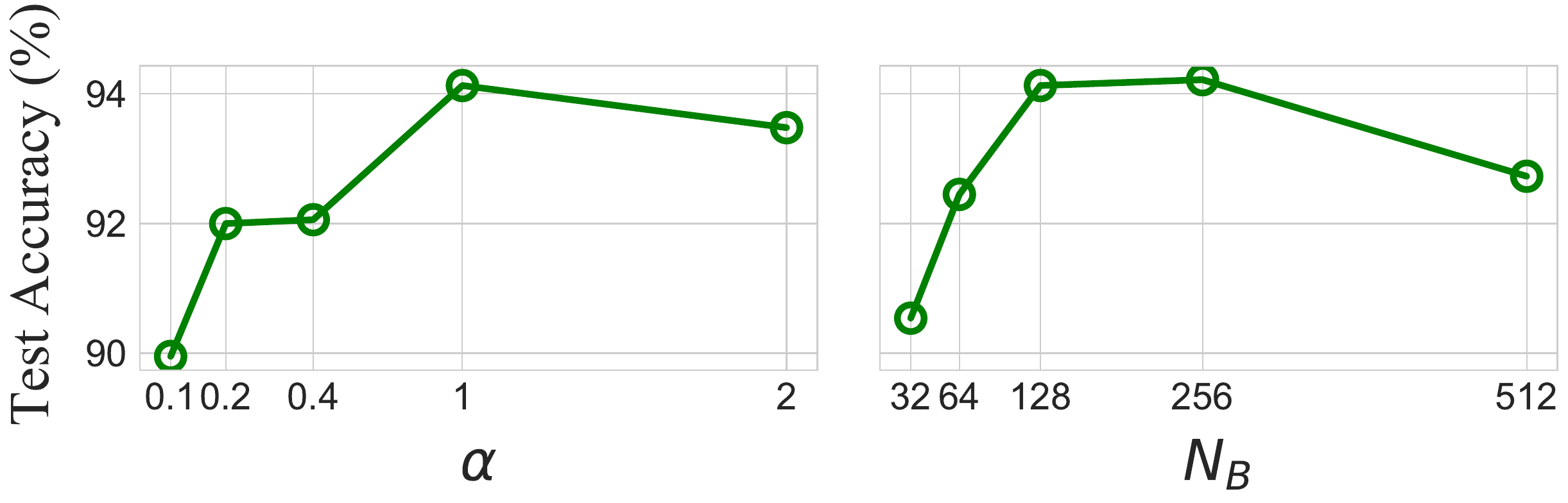}
  \caption{Ablation studies on invariance coefficient $\alpha$ and tracked batch number $N_{B}$. The experiments are conducted on CIFAR-10 under CADR's protocol with $\gamma=20$. }
  \label{fig:ab}  
\end{figure}

\noindent\textbf{Hyper-parameters.}  We investigate the effect of the class invariance coefficient $\alpha$ and the tracked batch number $N_{B}$ on PRG, which is shown in Fig. \ref{fig:ab}. Choosing an appropriate $\alpha$ to control the degree of class invariance in pseudo-rectifying is important for PRG, which ensures stability of supervision information and training. Meanwhile, we note that too small $N_{B}$ is not sufficient to estimate the underlying distribution of class transitions, where $N_{B}=128$ is a sensible choice for both memory overhead and performance.

\begin{table}
        \caption{Ablation studies on step $k$ of driving possible class transitions. We respectively report accuracy (\%) and GM on CIFAR-10 under CADR's protocol with $\gamma=20$.
  }
  
  \label{tab:ab2}
    \centering
    \footnotesize
    \setlength{\tabcolsep}{4.2mm}{
\begin{tabular}{@{}lcccc@{}}
\toprule
$k$ & $1$ (default PRG) &  $2$ & $5$  & $10$  \\ \midrule
Accuracy     &  94.04 & 91.33 & 87.76 & 82.60                   \\
GM&93.53 &90.79& 85.80& 80.24\\ \bottomrule
\end{tabular}
}
     \end{table}

\section{Conclusion}
This paper can be concluded as proposing a effective SSL framework called class transition based Pseudo-Rectifying Guidance (PRG) to address SSL in the MNAR scenarios. Firstly, we argue that the history of class transition caused by pseudo-rectifying can be utilized to offer informative guidance for future label assignment. Thus, we model the class transition as a Markov random walk along the nodes of the graph constructed on the class tracking matrix. Finally, we propose to utilize the class prediction information at current epoch (or last epoch) to guide the class transition for pseudo-rectifying so that the bias of label imputation can be alleviated. Except for MNAR, we believe PRG can be used for robust semi-supervised learning in broader scenarios.

\section*{Acknowledgement}
Yue Duan and Yinghuan Shi are with the National Key Laboratory for Novel Software Technology and the National Institute of Healthcare Data Science, Nanjing University. Lei Qi is with the School of Computer Science and Engineering, Southeast University. This work is supported by the NSFC Program (62222604, 62206052, 62192783), China Postdoctoral Science Foundation Project (2023T160100), Jiangsu Natural Science Foundation Project (BK20210224), and CCF-Lenovo Bule Ocean Research Fund.

{\small
\bibliographystyle{ieee_fullname}
\bibliography{egpaper_final}
}

\clearpage
\onecolumn
\appendix

\newcommand{\custommaketitle}[2]{%
   \newpage
   \null
   \vskip .375in
   \begin{center}
      {\Large \bfseries #1 \par}
      \vspace*{24pt}
      {\large
      \lineskip .5em
      \begin{tabular}[t]{c}
         #2
      \end{tabular}
      \par}
      \vskip .5em
      \vspace*{12pt}
   \end{center}
}

\custommaketitle{Towards Semi-supervised Learning with Non-random Missing Labels\\
- Supplementary Material -}{}
\vspace{-2em}
\thispagestyle{empty}
\section{Algorithm of PRG$^{\mathrm{Last}}$}
\label{sec:alg}
\begin{algorithm}[h]
    \small
      \caption{PRG$^{\mathrm{Last}}$: PRG Using Class Predictions from the Last Epoch} %
      \LinesNumbered 
      \label{a2}
      \KwIn{class tracking matrices $\mathcal{C}=\{ \mathbf{C}^{(i)};i\in (1,...,N_{B})\}$, labeled training dataset $D_{L}$, unlabeled training dataset $D_{U}$, model $\theta$, label bank $\{l^{(i)};i\in (1,...,n_{T}-n_{L})\}$}
      \For{$n=1$ \rm{\textbf{to}} $\mathrm{MaxIteration}$}{
      From $D_{L}$, draw a mini-batch $\mathcal{B}_{L}=\{(x^{(b)}_{L},y^{(b)}_{L});b\in (1,...,B)\}$\\
      From $D_{U}$, draw a mini-batch $\mathcal{B}_{U}=\{(x^{(b)}_{U});b\in (1,...,B_{U})\}$\\
      
  $\mathbf{H}=\mathrm{RowWiseNormalize}(\mathrm{Average}(\mathcal{C}))$ \hfill \tcp{Construct transition matrix} 
     $H_{ij}'= \frac{H_{ij}}{\frac{L_{j}}{\sum^{k}_{d=1}L_{d}}}$ \hfill\tcp{Rescale $\mathbf{H}$ at class-level}
  \For{$b=1$ \rm{\textbf{to}} $B_{U}$}{
            $p^{(b)}=f_{\theta}(x_{U}^{(b)})$ \hfill\tcp{Compute model prediction}
            $\mathrm{idx}=\mathrm{Index}(x^{(b)}_{U})$ \hfill\tcp{Obtain the index of $x^{(b)}_{U}$ in $D_{U}$}
            $\tilde{p}^{(b)} =\mathrm{Normalize}( H_{l^{(\mathrm{idx})}}' \circ p^{(b)})$\hfill\tcp{Perform pseudo-rectifying guidance}
            $\hat{p}^{(b)}=\arg\max(p^{(b)})$ \hfill\tcp{Compute class prediction}
            \If{$l^{(\mathrm{idx})}\neq \hat{p}^{(b)}$}{ $C^{(n)}_{l^{(\mathrm{idx})}\hat{p}^{(b)}}=C^{(n)}_{l^{(\mathrm{idx})}\hat{p}^{(b)}}+1$ \hfill\tcp{Perform class transition tracking}
            $l^{(\mathrm{idx})}=\hat{p}^{(b)}$
            }
            
  }
  $\mathcal{L}_{L},\mathcal{L}_{U}=\mathrm{FixMatch}\left(\mathcal{B}_{L},\mathcal{B}_{U},\{\tilde{p}^{(b)};b\in (1,...,B_{U})\}\right)$ \hfill\tcp{Run an applicable SSL learner}
         $\theta=\mathrm{SGD}(\mathcal{L}_{L}+\mathcal{L}_{U},\theta)$ \hfill\tcp{Update model parameters $\theta$}
      }
\end{algorithm}

\section{Discussion on Re-Weighting Scheme of $\mathbf{H}$}
\label{sec:appd}
In this section, we give insights into re-weighting scheme of $\mathbf{H}$ in Eq. (\ref{eq:sca}) based on the following theoretical justification. Overall, we give an explanation from the perspective of gradient. Our re-weighting scheme potentially scale the gradient magnitude on the learning of the unlabeled data to mitigate adverse effects of biased labeled data.
Letting $p$ be the naive soft label vector, by Eq. (\ref{eq:sca}), we re-weight $\mathbf{H}$ by $H_{ij}'= \times\frac{H_{ij}}{\frac{L_{j}}{\sum^{k}_{d=1}L_{d}}}$ and obtain the rescaled pseudo-label vector $\tilde{p}^{\mathtt{}}=\mathrm{Normalize}(\mathbf{H}'\circ p)$. Hence, the cross-entropy between prediction $p$ and $\tilde{p}^{\mathtt{}}$ can be formalized as 
\begin{align}
    \mathcal{L}_{U}=-\sum_{c}^{k}\tilde{p}^{\mathtt{}}\log{p_{c}}&=-\sum_{c}^{k}\left(\frac{H'_{ij}\times p_{c}}{\mathcal{Z}}\right)\log{p_{c}}\nonumber\\
    &=-\sum_{c}^{k}\left(\frac{H_{\hat{p}c}\times p_{c}}{\mathcal{Z}\frac{L_{c} }{\sum^{k}_{d=1}L_{d} }}\right)\log{p_{c}},
\end{align} 
where $\mathcal{Z}$ is the normalize factor.  $\frac{L_{c} }{\sum^{k}_{d=1}L_{d} }$ can be regarded as the ratio of pseudo-labels belonging to class $c$ to all labels. Denoting the logit outputted from the model as $o$ (implying $p=\mathrm{Softmax}(o)$), with no gradient on pseudo-label $\tilde{p}$, we obtain $\frac{\partial \mathcal{L}_{U}}{\partial o_{c}}=-\sum^{k}_{c}\frac{\tilde{p}_{c}}{p_{c}}\frac{\partial p_{c}}{\partial o_{c}}$, \ie,
\begin{align}
   \frac{\partial \mathcal{L}_{U}}{\partial o_{c}}&=-(\tilde{p}_{c}-\tilde{p}_{c}p_{c}-\sum^{k}_{i\neq c}\tilde{p}_{i}p_{c})\\
   &=\left(1-\frac{H_{\hat{p}c}}{\mathcal{Z}\frac{L_{c} }{\sum^{k}_{d=1}L_{d} }}\right)p_{c}. 
\end{align}
The larger the difference between $H_{\hat{p}c}$ and $\frac{L_{c} }{\sum^{k}_{d=1}L_{d} }$, the larger the gradient; and the smaller the difference between $H_{\hat{p}c}$ and $\frac{L_{c} }{\sum^{k}_{d=1}L_{d} }$, the smaller the gradient ($\frac{\partial \mathcal{L}_{U}}{\partial o_{c}}=0$ when $\frac{H_{\hat{p}c}}{\mathcal{Z}\frac{L_{c} }{\sum^{k}_{d=1}L_{d} }}=1$). This means that we intend to provide unbiased guidance (because this is derived from the unlabeled data) for the learning of unlabeled samples from the class level, so as to resist the influence of biased labeled samples.
In addition, the idea behind this re-weighting scheme is that 
the model should increase the learning effort for rare classes (the less labels a class is assigned, the smaller the $\frac{L_{c} }{\sum^{k}_{d=1}L_{d}}$, the larger the gradient) rather than overlearn popular classes. This will implicitly lead to the model not carrying out too many pseudo-rectifying processes resulting in more labels transition to classes with too many labels assigned, but trying to assign labels to rare classes.

\section{Implementation Details}
\label{sec:id2}
In this section, we show the complete hyper-parameters in Tab. \ref{hyper}. As mentioned in Sec. 4, our method is implemented as a plugin to FixMatch \cite{sohn2020fixmatch}. Thus, we keep the original hyper-parameters in FixMatch and alert additional hyper-parameters in our method. Note that FixMatch sets different values of weight decay $w$ for CIFAR-10 and CIFAR-100, which are 0.0005 and 0.001 respectively. For simplicity, we set $w=0.0005$ for all experiments in our work. Additionally, 
the models in this paper are trained on GeForce RTX 3090/2080 Ti and Tesla V100. We observe that since no additional network components are introduced, the average running time of single iteration hardly increased, which means our method does not introduce excessive computational overhead.

\begin{table}[h]
  \centering
  \footnotesize
  \caption{Complete list of hyper-parameters of PRG plugged in FixMatch \cite{sohn2020fixmatch}. $N_{B}$ and $\alpha$ are additional hyper-parameters in our method whereas other hyper-parameters follow the setting of original FixMatch. Note that unlabeled data batch size $B_{U}$ can be calculated by $B_{U}=\mu B$.} 
  \label{hyper}
 \setlength{\tabcolsep}{5.7mm}{
  \begin{tabular}{@{}l|l|c|c|c@{}}
  \toprule
  Hyper-parameter                 & Description        & CIFAR-10             &CIFAR-100 &mini-ImageNet               \\\midrule
  $\mu$            & The ratio of unlabeled data to labeled data in a mini-batch             &\multicolumn{3}{c}{7}    \\
  $B$            &  Batch size for labeled data and class transition tracking          &\multicolumn{3}{c}{64}    \\ 
  $B_{U}$            &  Batch size for unlabeled data          &\multicolumn{3}{c}{448}    \\ 
  $\lambda_{U}$          & Unlabeled loss weight          &\multicolumn{3}{c}{1}     \\
    $\tau$            & Confidence threshold         &\multicolumn{3}{c}{0.95}   \\ 
  $lr$           & Start learning rate           &\multicolumn{3}{c}{0.03}    \\
  $\beta$            & Momentum          &\multicolumn{3}{c}{0.9}    \\
  $w$        & Weight decay           &\multicolumn{3}{c}{0.0005}    \\
  $N_{B}$        & Tracked batch number           &\multicolumn{3}{c}{128}    \\
  $\alpha$        & Class invariance coefficient        &\multicolumn{3}{c}{1}\\\bottomrule
  \end{tabular}
  }
\end{table}

\section{Additional Experimental Results}
\subsection{Using Distribution Alignment in MNAR}
\label{sec:dainm}
As discussed in Sec. \ref{sec:ctt}, \textit{distribution alignment} (DA) aims to perform strong regularization on pseudo-labels by aligning the class distribution of predictions on unlabeled data to that of labeled data. DA boosts the performance of SSL models tangibly \cite{berthelot2020remixmatch,gong2021alphamatch,li2021comatch,sohn2020fixmatch}. However, DA works on a strong assumption that the distribution of unlabeled data matches that of labeled data. In MNAR, this assumption does not hold obviously. Therefore, SSL methods that incorporate DA will face predicaments in MNAR.
As shown in Tab. \ref{tab:app}, rather than improving performance, integrating DA into SSL models is counterproductive, \eg, original FixMatch outperforms FixMatch with DA by up to 28.68\% on CIFAR-10. 
Another example is SimMatch in Tab. \ref{tab:cadr}. Despite SimMatch being a considerably more advanced method compared to FixMatch, its performance with PRG is weaker than that of FixMatch when a small value of $\gamma$ is used, implying a small $n_L$. This underperformance can be attributed to its adoption of DA. As $\gamma$ (implying $n_L$) increases, more supervisory information allows SimMatch's inherent strong performance begins to overshadow the negative impact of DA.
Conversely, our method is not restricted by the mismatched distributions and achieves superior performance across the board, because PRG helps the model to better handle MNAR scenarios without any prior information (distribution prior estimated from labeled data is used in  DA). 
\begin{table}[h]

  \footnotesize
  \caption{Accuracy (\%) in MNAR under our protocol compared with more baseline methods using distribution alignment ({DA}) \cite{berthelot2020remixmatch}. Note that CoMatch \cite{li2021comatch} (a recently-proposed graph-based SSL method integrating contrastive learning) also combines DA to improve the quality of pseudo-labels in the conventional SSL setting. 
 }
  \label{tab:app}
  \vskip 0in
    \centering
    
    \setlength{\tabcolsep}{2.75mm}{
      \begin{tabular}{@{}l|cccc|cc|cc@{}}     
      \toprule
      \multirow{3}{*}{Method}& \multicolumn{2}{c}{CIFAR-10 ($n_{L}$ = 40)} & \multicolumn{2}{c|}{CIFAR-10 ($n_{L}$ = 250)} & \multicolumn{2}{c|}{CIFAR-100 ($n_{L}$ = 2500)} & \multicolumn{2}{c}{mini-ImageNet ($n_{L}$ = 1000)}  \\  \cmidrule(lr){2-3} \cmidrule(lr){4-5}  \cmidrule(lr){6-7}  \cmidrule(lr){8-9} 
   
      & $N_{1}$ = 10            & 20                       &  100 & 200 &  ~~~~100 & 200      & ~~~~40& 80   \\ \cmidrule(r){1-1}\cmidrule(lr){2-3} \cmidrule(lr){4-5}  \cmidrule(lr){6-7}  \cmidrule(lr){8-9}    
      CoMatch    & 60.27& 39.48          & 57.87&26.77&~~~~48.02&30.08&~~~~30.24&21.47 \\
     \cmidrule(r){1-1} \cmidrule(lr){2-3} \cmidrule(lr){4-5}  \cmidrule(lr){6-7}  \cmidrule(lr){8-9}  
      FixMatch    & 85.72 &76.53     & 69.76&46.53&~~~~{61.31}&41.38&~~~~36.20&28.33 \\
      + DA  &71.23\tolr{14.49} & 47.85\tolr{28.68} & 61.8\tolr{7.96} & 27.61\tolr{18.92} & 50.94\tolr{10.37} & 31.82\tolr{9.56} & 33.87\tolr{2.33} & 23.53\tolr{4.78} \\
      + PRG (Ours)   & \textbf{91.87\tolb{6.15}} & 77.44\tolb{0.91} & \textbf{93.93\tolb{24.17}} & \textbf{67.86\tolb{21.33}} & \textbf{61.49\tolb{0.18}} & \textbf{49.84\tolb{8.46}} & \textbf{39.99\tolb{3.79}} & \textbf{35.39\tolb{7.069}} \\
      + PRG$^{\mathrm{Last}}$ (Ours)  & 85.66\tolr{0.06} & \textbf{77.85\tolb{1.32}} & 92.80\tolb{23.04} & 64.00\tolb{17.47} & 60.41\tolr{0.90} & 43.80\tolb{2.42} & 39.84\tolb{3.64} & 33.10\tolb{4.77} \\
    
      \bottomrule
      \end{tabular}
  }
\end{table}

\begin{figure}[t]
  \centering
  \resizebox{\linewidth}{!}{   \includegraphics[]{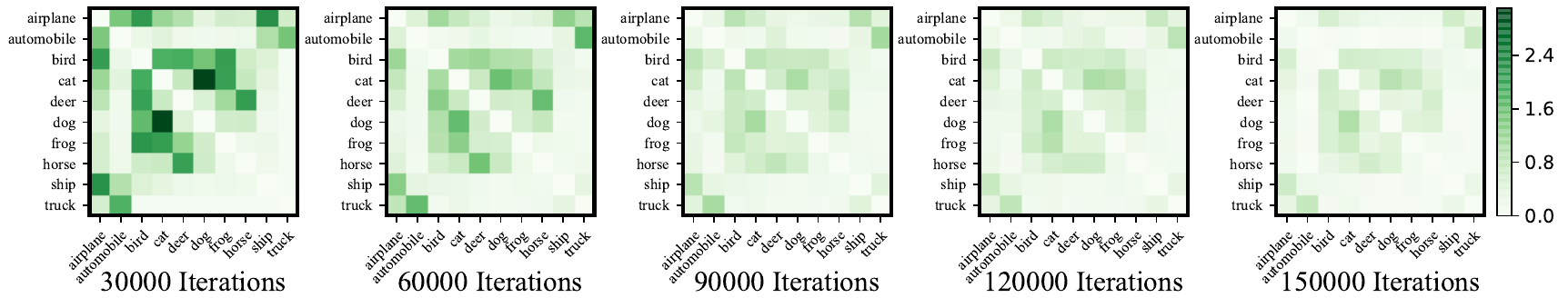}}
  \vspace{-1.5em}
  \caption{Visualization of class tracking matrix $\mathbf{C}$ obtained in training process of FixMatch \cite{sohn2020fixmatch} combining PRG. Experiments are conducted on CIFAR-10 with the same setting as in Fig. \ref{fig:met}. }
  \label{fig:met-ldg}
\end{figure}

\subsection{Empirical Analysis on PRG}
\label{sec:eaop}
Different from Fig. \ref{fig:met}, the color blocks in the heatmap in Fig. \ref{fig:met-ldg} almost cover the entire diagram, and some color blocks are not missing as the training progresses, \ie, with the help of PRG, the information exchange between classes remains frequent during the learning process, and the model maintains the pseudo-rectifying ability for almost all classes.

\subsection{More Evaluations on PRG}
\label{sec:moree}

\begin{wraptable}{r}{6.5cm}
\vskip -1.45em
\caption{Accuracy (\%) on CIFAR-10 with $n_{L}=40$ and various $\gamma_{u}$ under our protocol.}
  \vskip 0in
  \label{tab:ab44}
    \centering
    \footnotesize
     
\begin{tabular}{@{}lccc@{}}
\toprule
Method & $\gamma_u=20$ &  $\gamma_u=50$ & $\gamma_u=100$    \\ \midrule
CoMatch  &  52.73&46.20&38.85                  \\\midrule
FixMatch     & 57.54&54.82&50.67 \\
+ DA     & 54.08\tolr{3.46} & 46.71\tolr{8.11} & 41.37\tolr{9.30} \\
+ CADR & 49.38\tolr{8.16} & 45.27\tolr{9.55} & 42.30\tolr{8.37}  \\
+ PRG (Ours) & \textbf{62.43\tolb{4.90}} & \textbf{62.44\tolb{7.62}} & \textbf{58.23\tolb{7.56}} \\\bottomrule
\end{tabular}

\end{wraptable}
\subsubsection{More MNAR Scenarios}
\label{app:mns}

We also provide more experiments on the setting of balanced labeled data with imbalanced unlabeled data, which is summarized in Tab. \ref{tab:ab44}. For specific, we set $n_{L}=40$ with balanced distribution and set $\gamma_{u}=50,100,200$ for imbalanced unlabeled data, \ie, the class-wise number of unlabeled data $M_{i}=M_{1}\times \gamma_{u}^{-\frac{k-i}{k-1}}$, where $M_{1}=5000$ in CIFAR-10. As shown in Tab. \ref{tab:ab44}, PRG  outperforms all baseline methods by a large margin (the performance of CADR is even weaker than original FixMatch), proving the robustness of PRG in this MNAR scenario due to the unbiased guidance derived from the class transition history.

\subsubsection{More Metrics} 
\label{app:mm}
To comprehensively explore the improvement of PRG in MNAR, we report the difference in class-wise precision and recall  with/without PRG. The experimental results are shown in Tab. \ref{tab:pr}. Compared to original FixMatch, we witness FixMatch with PRG achieves higer precision/recall by and large, especially on rare classes (\ie, class with larger index), which demonstrates that the bias removal capability of PRG effectively mitigates the effect of MNAR on the model. We also observe that both PRG and FixMatch achieve high precision as well as recall on popular classes and high precision but low recall on rare classes (especially FixMatch) in the early training period. The improvement of recall by PRG is due to the activated class transitions, which gives the model a certain probability to assign pseudo-labels to rare classes. 

\begin{table}[t]
\caption{Class-wise precision and recall on CIFAR-10 during the training under CADR’s protocol with $\gamma=50$.
  }
  \label{tab:pr}
    \centering
    \footnotesize
\begin{tabular}{@{}|c|c|cc|cc|cc|@{}}
\hline
\multirow{2}{*}{Method} & \multirow{2}{*}{Class Index} & \multicolumn{2}{c|}{30000 Iterations} &  \multicolumn{2}{c|}{90000 Iterations} & \multicolumn{2}{c|}{150000 Iterations}  \\ \cline{3-8}
&&Precision & Recall& Precision & Recall& Precision & Recall \\\hline
\multirow{10}{*}{FixMatch}&1&	45.21 & 95.22	&46.89 & 96.72	&47.93 & 97.80\\
&2&	49.12& 99.01&	49.59& 99.27&	50.27& 98.72\\
&3&	38.49& 88.73&	39.74& 88.43&	70.26& 89.47\\
&4&	75.02& 68.13&	75.63& 72.19&	82.04& 75.93\\
&5&	86.14& 88.43&	86.88& 90.21 &	88.42  & 94.38\\
&6&	89.45& 62.93&	91.03& 64.4&	89.31& 75.98\\
&7&	86.47& 90.03&	90.23& 8.89  &	91.37& 94.80\\
&8&	89.09& 75.94&	90.48& 75.21&	95.32 &  75.37\\
&9&	99.02& 0.00&	97.95& 1.00&	97.21& 2.00\\
&10&	0.00& 0.00 &99.60& 0.33&	98.60& 0.67\\\hline
\multirow{10}{*}{+ PRG (Ours)}&1&	70.52& 93.52&	87.34& 95.50&	88.25& 95.34\\
&2&	82.53& 98.21&	96.03& 98.32&	96.78& 98.56\\
&3&	73.52& 76.54&	90.92& 89.85&	92.37& 90.57\\
&4&	70.21& 73.77&	85.36  & 80.51&	87.89 & 81.37\\
&5&	79.03& 86.57&	90.31& 96.31&	92.74& 96.19\\
&6&	74.55& 61.03&	90.58& 79.88&	90.97& 82.43\\
&7&	89.12& 91.40&	93.09& 97.02&	93.79& 98.03\\
&8&	92.58& 80.14&	95.01& 96.21 &	96.32& 97.50\\
&9&	96.31& 76.50 &	95.22& 92.12&	95.63& 93.55\\
&10&	96.56& 62.52&	96.95& 96.01&	97.15& 96.81\\ \hline
\end{tabular}
\end{table}

\subsubsection{More SSL Learners} 
\label{app:msl}
Moreover, to further evaluate PRG's performance, we consider building PRG on the top of more SSL frameworks. Thus, we firstly conduct experiments on CIFAR-10 under CADR’s protocol with UPS \cite{rizve2021in} combining PRG. UPS is a recently-proposed uncertainty-aware pseudo-label selection framework for SSL, which is the SOTA method among pseudo-labeling based methods. We keep all training settings the same as the original UPS. With $\gamma=20$, UPS achieves an accuracy of \textbf{30.46\%} whereas UPS with PRG achieves an accuracy of \textbf{32.22\%}. We note that UPS performs poorly in the MNAR scenarios because it is a more pure pseudo-labeling approach that does not introduce consistency regularization to improve model performance. Also we observe that PRG improves UPS marginally, much less than FixMatch. This is understandable because the negative learning that UPS prides itself on can be potentially negatively affected by the probability distribution of pseudo-label being adjusted by PRG, \eg, uncertainty being altered. Next, we adopt a more advanced SSL learner FlexMatch \cite{zhang2021flexmatch} to evaluate PRG, which is shown in Tab. \ref{tab:app2}. PRG still complements the unrobustness of this strong SSL method in MNAR.
\begin{table}[t]

  \footnotesize
  \caption{Accuracy (\%) in MNAR under our protocol with more SSL learners. 
 }
  \label{tab:app2}
  \vskip 0in
    \centering
    
    \setlength{\tabcolsep}{2.95mm}{
      \begin{tabular}{@{}l|cccc|cc|cc@{}}     
      \toprule
      \multirow{3}{*}{Method}& \multicolumn{2}{c}{CIFAR-10 ($n_{L}$ = 40)} & \multicolumn{2}{c|}{CIFAR-10 ($n_{L}$ = 250)} & \multicolumn{2}{c|}{CIFAR-100 ($n_{L}$ = 2500)} & \multicolumn{2}{c}{mini-ImageNet ($n_{L}$ = 1000)}  \\  \cmidrule(lr){2-3} \cmidrule(lr){4-5}  \cmidrule(lr){6-7}  \cmidrule(lr){8-9} 
   
      & $N_{1}$ = 10            & 20                       &  100 & 200 &  ~~~~100 & 200      & ~~~~40& 80   \\ \cmidrule(r){1-1}\cmidrule(lr){2-3} \cmidrule(lr){4-5}  \cmidrule(lr){6-7}  \cmidrule(lr){8-9}     
      FlexMatch    & 90.86 &84.53    & 79.13&55.40&61.49&45.26&39.45&34.18 \\
      + PRG (Ours)   & \textbf{92.17}\tolb{1.31}    & {88.46}\tolb{3.93}      & \textbf{93.95}\tolb{14.82}     &  \textbf{69.88}\tolb{14.48}      & \textbf{65.29}\tolb{3.80}            & \textbf{50.31}\tolb{5.05}        & {41.02}\tolb{1.57}      & \textbf{36.59}\tolb{2.41}      \\ 
      + PRG$^{\mathrm{Last}}$ (Ours)  & 91.03\tolb{0.17} & \textbf{89.42\tolb{4.89}} & 92.94\tolb{13.81} & {67.07\tolb{11.67}} & 64.66\tolr{3.17} & 48.82\tolb{3.56} & \textbf{41.25\tolb{1.80}} & 35.16\tolb{0.98} \\
     
      \bottomrule
      \end{tabular}
  }
\end{table}
\begin{table*}[t]
  \footnotesize
  \caption{Accuracy (\%) in the conventional setting with various $n_{L}$. Results of baselines are reported in CADR \cite{hu2021non} while results of $^{\ast}$ are based on our reimplementation.}
  \label{tab:con}
  \vskip -0em
    \centering
  \setlength{\tabcolsep}{3.8mm}{
      \begin{tabular}{@{}l|ccc|ccc|c@{}}     
      \toprule
      \multirow{3}{*}{Method}  & \multicolumn{3}{c|}{CIFAR-10} & \multicolumn{3}{c|}{CIFAR-100} & \multicolumn{1}{c}{mini-ImageNet}  \\ \cmidrule(lr){2-4} \cmidrule(lr){5-7}  \cmidrule(lr){8-8}  
   
                      & $n_{L}$ = 40  & 250 &4000  &  400  & 2500      & 10000  &1000    \\ \cmidrule(r){1-1} \cmidrule(lr){2-4} \cmidrule(lr){5-7}  \cmidrule(lr){8-8}  
      FixMatch      &88.61\tols{3.35}&\textbf{94.93\tols{0.33}}&95.69\tols{0.15}&50.05\tols{3.01}&\textbf{71.36\tols{0.24}}&76.82\tols{0.11}&$\mbox{39.03\tols{0.66}}^\ast$ \\
      + CADR & 94.41\tolb{5.80}  &94.35\tolr{0.58} &95.59\tolr{0.10} &\textbf{52.90\tolb{2.85}}&70.61\tolr{0.75} &76.93\tolb{0.11}  &-\\
      + PRG (Ours)       & \textbf{94.44\tol{0.16}{5.83}} & 94.42\toll{0.06}{0.51}      &  95.38\toll{0.10}{0.31}    & 52.45\tol{3.75}{2.40}         & 70.12\toll{0.21}{1.24}     & 76.49\toll{0.42}{0.33}  & 47.34\tol{1.60}{8.31}   \\
      + PRG$^{\mathrm{Last}}$ (Ours)        & 93.00\tol{0.79}{4.39}  & {94.43}\toll{0.33}{0.50}   &  \textbf{95.75\tol{0.11}{0.06}}      & {48.81\toll{0.15}{1.24}}       & {70.01\toll{0.02}{1.35}}      & \textbf{77.12\tol{0.13}{0.30}}   & \textbf{48.23\tol{0.59}{9.20}}      \\ %
      \bottomrule
      \end{tabular}
}
\vskip -0em
\end{table*}
\subsubsection{More Data Types}
\label{app:mdt}
The results of VIME combined with PRG on tabular data  are shown in Tab. \ref{tab:ab11}. VIME \cite{yoon2020vime} is a prevailing self- and semi-supervised learning frameworks for tabular data with pretext task of estimating mask vectors from corrupted tabular data. We implement PRG above the semi-supervised learning component of VIME. PRG provide pseudo-rectifying guidance to rescale the pseudo-labels for  the original unlabeled sample in VIME. Specially, we replace the consistency loss used in VIME (\ie, mean squared error in Eq. (9) in \cite{yoon2020vime}) with standard cross-entropy loss to makes PRG applicable to VIME. 

\subsubsection{Coventional SSL Setting}
\label{app:con}
As shown in Tab. \ref{tab:con}, our method still works well in the conventional SSL setting, \ie, both the labeled data and the unlabeled data are  \textit{balanced}. The class-level guidance offered by our method is also valid in the conventional setting while maintaining the vitality of class transition, even though there is not too much need to remove bias on label imputation.

\end{document}